\pdfoutput=1

\documentclass[11pt]{article}

\usepackage[]{acl}

\usepackage{times}
\usepackage{latexsym}
\usepackage{enumitem}
\usepackage{multirow}
\usepackage{amsmath}
\usepackage{booktabs}
\usepackage{subfigure}
\usepackage{diagbox}
\newcommand{\modi}{\textcolor{black}}

\usepackage[T1]{fontenc}

\usepackage[utf8]{inputenc}

\usepackage{microtype}

\usepackage{graphicx}

\title{An Empirical Study of Many-to-Many Summarization with Large Language Models}

\author{Jiaan Wang\textsuperscript{1}, \ Fandong Meng\textsuperscript{1}\thanks{ \ \ Corresponding author.}, \  Zengkui Sun\textsuperscript{4}, \ Yunlong Liang\textsuperscript{1}, \ Yuxuan Cao\textsuperscript{5} \\
\bf {Jiarong Xu\textsuperscript{2}, \ Haoxiang Shi\textsuperscript{3} and Jie Zhou\textsuperscript{1} } \\
\textsuperscript{1}Pattern Recognition Center, WeChat AI, Tencent Inc, China \quad \textsuperscript{2}Fudan Unversity \\
\textsuperscript{3}Waseda University \quad \textsuperscript{4}Beijing Jiaotong University \quad \textsuperscript{5}Zhejiang University \\
\texttt{\{torchwang,fandongmeng,withtomzhou\}@tencent.com}
}

\begin{document}
\maketitle
\begin{abstract}
Many-to-many summarization (M2MS) aims to process documents in any language and generate the corresponding summaries also in any language.
Recently, large language models (LLMs) have shown strong multi-lingual abilities, giving them the potential to perform M2MS in real applications.
\modi{This work presents a systematic empirical study on LLMs' M2MS ability.
Specifically, we first reorganize M2MS data based on eight previous domain-specific datasets.
The reorganized data contains 47.8K samples spanning five domains and six languages, which could be used to train and evaluate LLMs.}
Then, we benchmark \modi{18} LLMs in a zero-shot manner and an instruction-tuning manner.
Fine-tuned traditional models (\emph{e.g.}, mBART) are also conducted for comparisons.
Our experiments reveal that, zero-shot LLMs achieve competitive results with fine-tuned traditional models.
After instruct-tuning, open-source LLMs can significantly improve their M2MS ability, and outperform zero-shot LLMs (including GPT-4) in terms of automatic evaluations.
In addition, we demonstrate this task-specific improvement does not sacrifice the LLMs' general task-solving abilities.
\modi{However, as revealed by our human evaluation, LLMs still face the factuality issue, and the instruction tuning might intensify the issue.
Thus, how to control factual errors becomes the key when building LLM summarizers in real applications, and is worthy to be noted in future research.}

\end{abstract}

\begin{figure}[t]
\centerline{\includegraphics[width=0.35\textwidth]{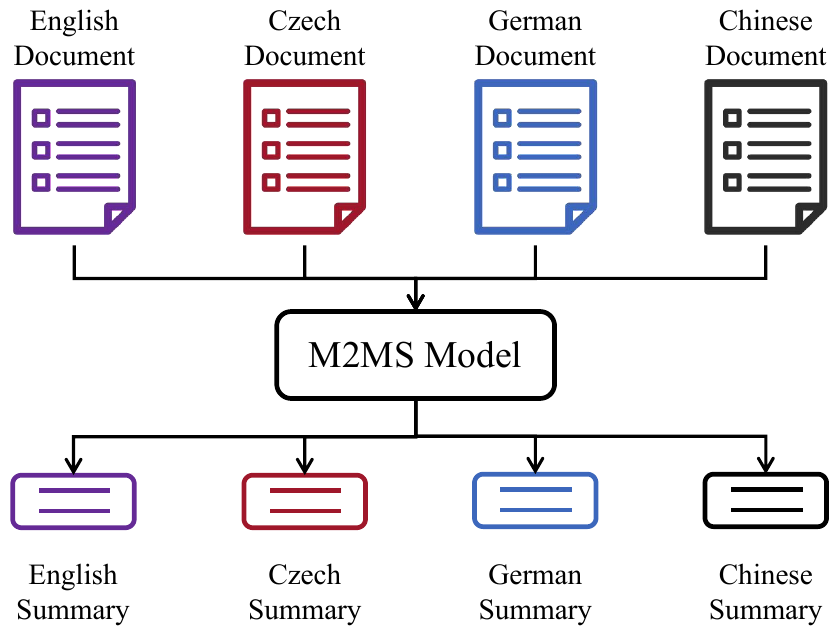}}
\caption{Illustration of many-to-many summarization.}
\label{fig:intro}
\end{figure}

\section{Introduction}
Many-to-Many Summarization (M2MS) is proposed to generate a brief summary in any language given a document also in any language (c.f., Figure~\ref{fig:intro})~\cite{wang-etal-2023-towards-unifying,bhattacharjee-etal-2023-crosssum}.
This task is extremely challenging since it requires the ability to summarize and translate across many languages.
Meanwhile, many LLMs adopt the multi-lingual setting to share the language modeling across various languages~\cite{touvron2023llama,OpenAI2023GPT4TR}, making it possible to become an advanced M2MS solver in theory.
However, there is still a lack of practice in exploring LLMs' M2MS performance.

In this paper, we try to investigate how LLMs can perform the M2MS task in real applications with multi-domain scenarios.
Considering the limited diversity and the single-domain characteristic in each existing dataset, a single dataset cannot be directly used to benchmark LLMs in multi-domain scenarios. Thus, we first reorganize and select M2MS samples from eight existing multi-lingual summarization datasets~\cite{ladhak-etal-2020-wikilingua,fatima-strube-2021-novel,perez-beltrachini-lapata-2021-models,wang-etal-2022-clidsum,chen-etal-2023-revisiting,bhattacharjee-etal-2023-crosssum,zheng2023long}.
These datasets cover five domains, \emph{i.e.}, news, how-to guides, encyclopedia, dialogue, and technology, allowing the transformation of shareable knowledge across different domains.
During sample selection, we consider the intrinsic quality metrics (coverage, redundancy and coherence), and balance the number of samples among different languages and domains.
\modi{For the testing samples, we also consider data contamination to ensure fair evaluations.}
After that, there are 47.8K samples used to train and evaluate models in our study.

Then, we benchmark \modi{18} human-aligned LLMs, including open- and closed-source LLMs. We evaluate their zero-shot M2MS ability by prompting them with task-specific instruction and in-context examples. In this way, LLMs can leverage their instruction-following ability to perform M2MS without any parameter updating.
Besides, we fine-tune two state-of-the-art traditional models~\cite{tang-etal-2021-multilingual,wang-etal-2023-towards-unifying} for comparison.
Our experiments reveal that the zero-shot LLMs could achieve competitive results with fine-tuned traditional models, showing the promising task-solving ability of LLMs.
Furthermore, we train the open-source LLMs to perform M2MS via instruction tuning.
We find that open-source LLMs can significantly improve their M2MS ability through instruct-tuning, and outperform the original models and traditional models by a large margin.
Some tuned LLMs can even outperform zero-shot GPT-4 in terms of automatic metrics.
In addition, we evaluate the original LLMs and the instruction-tuned LLMs on MMLU~\cite{hendrycks2021measuring}.
The results demonstrate the improvement brought by instruction-tuning does not sacrifice the LLMs' general task-solving abilities.

\modi{Moreover, as revealed by recent work, hallucination is an obstacle when building LLMs in real applications~\cite{zhang2023siren}.
We conduct fine-grained human evaluation to figure out whether the generated summaries involve factual errors.
The results indicate that open-source LLMs have more factual errors than zero-shot GPT-4.
Besides, the instruction tuning on LLMs might intensify factual errors, and make LLMs tend to generate hallucinations.
This issue might come from the hallucination signals of ground truth references in existing summarization datasets~\cite{wang-etal-2022-analyzing,gao-etal-2023-evaluating}.
Therefore, future work should strengthen the factual consistency when building LLM summarizers in real M2MS applications.}

Our main contributions are concluded as follows:
\begin{itemize}[leftmargin=*,topsep=0pt]
\setlength{\itemsep}{0pt}
\setlength{\parsep}{0pt}
\setlength{\parskip}{0pt}
\item To our knowledge, we are the first to investigate how LLMs perform the M2MS task. To this end, we reorganize and select samples from previous multi-lingual summarization datasets to construct a multi-domain M2MS scenario.

\item We conduct extensive studies on \modi{18} LLMs. The evaluation process involves zero-shot prompting and instruction tuning. Fine-tuned traditional models are also conducted for comparison.

\item In-depth analyses of the M2MS results on automatic evaluation and human evaluation provide a deeper understanding of the M2MS task-solving situations in the LLM era.

\end{itemize}

\section{Related Work}

\subsection{Summarization in Multi-Lingual World.}

To adapt text summarization to the multilingual world, the summarization research field proposes the following three branch tasks:

(1) Cross-lingual summarization (CLS) aims to generate a target-language summary for a document in a \emph{different} source language~\cite{wang-etal-2022-survey}.
Early work typically focuses on pipeline methods~\cite{Leuski2003CrosslingualCE,Orasan2008EvaluationOA}.
Some recent studies have demonstrated that such pipeline methods suffer from error propagation and inference latency, and their performance is worse than the end-to-end ones~\cite{zhu-etal-2019-ncls,perez-beltrachini-lapata-2021-models}.
Meanwhile, with the availability of large-scale CLS datasets,
many researchers shift the research attention to end-to-end CLS, using different techniques to deal with CLS, \emph{i.e.}, multi-task learning~\cite{cao-etal-2020-jointly,bai2022unifying,Liang2022AVH}, knowledge distillation~\cite{duan-etal-2019-zero,Nguyen2021ImprovingNC} and different pre-training strategies~\cite{xu-etal-2020-mixed,chi-etal-2021-mt6,wang-etal-2022-clidsum}.

(2) Multi-lingual summarization (MLS) aims to process documents in multiple languages and generate their summaries in the \emph{corresponding} language.
Recently, large-scale MLS datasets~\cite{scialom-etal-2020-mlsum,hasan-etal-2021-xl,liang-etal-2023-summary} have been proposed one after another, facilitating further research on MLS.
MultiSumm~\cite{Cao_Wan_Yao_Yu_2020} explores various knowledge-sharing strategies to train a MLS model among different languages.
CALMS~\cite{wang-etal-2021-contrastive} proposes to train MLS models in the contrastive learning framework to share salient information extractive ability across different languages.
Some studies~\cite{aharoni-etal-2023-multilingual,qiu-etal-2023-detecting} aim to enhance the factual consistency of MLS models.

(3) Many-to-many summarization (M2MS) combines CLS and MLS into a more general setting that requires the model to summarize documents in any source language to a target language also in any language.
CrossSum~\cite{bhattacharjee-etal-2023-crosssum} studies M2MS in the news domain, and shows M2MS model consistently outperforms CLS models, verifying the practicality of M2MS.
\citet{wang-etal-2023-towards-unifying} propose PISCES, a pre-trained M2MS model based on mBART.
Besides, they explore the summarization models trained with the settings of CLS, MLS and M2MS, and demonstrates the superiority of M2MS that allows task knowledge sharing across all languages.
Different from existing work which generally uses traditional models, such as mBART~\cite{liu-etal-2020-multilingual-denoising},
we first explore how well existing LLMs can perform M2MS in the zero-shot and the instruction-tuning manners.

\subsection{Large Language Models.}

Recently, there has been growing interest in LLMs for various NLP tasks~\cite{zhao2023survey}.
A remarkable progress is the launch of ChatGPT~\cite{ChatGPT} and GPT-4~\cite{OpenAI2023GPT4TR}.
LLMs show their powerful ability that serve as a general-purpose language task solver.
Many powerful LLMs are proposed one after another to facilitate the LLM research, including LLaMa~\cite{touvron2023llama}, LLaMa-2~\cite{touvron2023llama2}, BaiChuan~\cite{yang2023baichuan}, Qwen~\cite{bai2023qwen,qwen2}, Vicuna~\cite{vicuna2023} and InternLM~\cite{team2023internlm}.
These LLMs generally adopt a three-stage training paradigm which first uses the next token prediction to learn the language modeling ability, and then leverages instruction tuning to enhance the model ability of following human instructions. Finally, an optional reinforcement learning with human feedback (RLHF) stage aligns LLMs' values with humans.

\section{Data}

\noindent \textbf{Dataset Selection.} 
To ensure the data quality, the selected multi-lingual summarization datasets should meet the following requirements:
(\romannumeral1) the datasets should be peer-reviewed and published;
(\romannumeral2) the datasets should provide cross-lingual alignments for their documents and summaries across different languages to support M2MS.\footnote{Note that not all datasets provide these alignments, some datasets only provide monolingual document-summary pairs in multiple languages, and thus do not support summarizing documents from a language into other languages.}
After carefully comparing existing data, we finally choose the following eight datasets:
(1) CrossSum~\cite{bhattacharjee-etal-2023-crosssum} is a news-domain dataset that collects document-summary pairs from the BBC news website.
(2) XWikis~\cite{perez-beltrachini-lapata-2021-models} is an encyclopedia-domain dataset that collects summarization samples from Wikipedia.
(3) XSAMSum~\cite{wang-etal-2022-clidsum}, (4) XMediaSum~\cite{wang-etal-2022-clidsum} and (5) DialogSumX~\cite{chen-etal-2023-revisiting} are three dialogue-domain datasets, which are collected by manually translating the summaries of existing English dialogue summarization datasets into other languages.
(6) WikiLingua~\cite{ladhak-etal-2020-wikilingua} is a multi-lingual dataset in the domain of how-to guides. This dataset is collected from the WikiHow website.
(7) Perseus~\cite{zheng2023long} is a technology-domain dataset that collects Chinese scientific articles with the corresponding Chinese and English summaries.
(8) Spektrum~\cite{fatima-strube-2021-novel} is also a technology-domain dataset. This dataset collects samples from Spektrum der Wissenschaf (a German scientific journal).

Considering the involved languages of the chosen datasets, we make the used data support English (abbr. En), Czech (Cs), German (De), French (Fr), Chinese (Zh) and Ukrainian (Uk).

\noindent \textbf{Intrinsic Metrics.} After determining the datasets and the languages, we select M2MS samples from these datasets to use in our empirical study.
Since the samples from a single dataset might be mixed-quality, we follow \citet{grusky-etal-2018-newsroom,bommasani-cardie-2020-intrinsic} and filter out low-quality samples based on three intrinsic quality metrics, \emph{i.e.}, coverage, redundancy and coherence.
These metrics are all automatically calculated based on the text features of document-summary pairs. For more details of these metrics and the filtering thresholds, please refer to Appendix~\ref{sec:appendix_intrinsic_metrics}.

\noindent \textbf{Size and \modi{Contamination} Controlling.} To ensure the data diversity in our empirical study, given a dataset and a specific source-target language pair, we decide to randomly select a few hundred samples from the remaining samples. Following the success of the instruction tuning in LLaMa-2~\cite{touvron2023llama2} and Vicuna~\cite{vicuna2023}, we control the number of training set to tens of thousands.
During sample selection, we also consider the balance of each language as well as each domain.
We make the selected data contain 19,530, \modi{14,150} and \modi{14,150} samples in the training, validation and testing sets.
\modi{As LLMs are pre-trained on massive data, their downstream performances might be inflated due to data contamination~\cite{dong-etal-2024-generalization,golchin2024time}.
To alleviate this issue, during the selection of the testing samples, we follow \citet{golchin2024time} to calculate instance-level contamination for each M2MS sample, and control the proportion of contaminated samples is less than 1\% in the testing set (more details are provided in Appendix~\ref{sec:appendix_data_contamination}).}

\noindent \textbf{Data Statistics.} As shown in Table~\ref{table:statistics1}, the final data covers most language pairs among the six languages except for Cs$\leftrightarrow$Uk and De$\leftrightarrow$Uk.
For more details, including the number of samples w.r.t each subset, the data sources w.r.t each language pair, length distribution and domain distribution, please refer to Appendix~\ref{sec:appendix_data_statistics}.

\begin{table}[t]
\centering
\resizebox{0.40\textwidth}{!}
{
\begin{tabular}{ccccccc}
\toprule[1pt]
 \diagbox[dir=NW]{Src}{Tgt}  & En                           & Cs                 & De                   & Fr               & Zh                       & Uk                 \\ \midrule[1pt]
En  &  3,900  & 1,200      & 3,400     & 1,350  & 3,050 & 1,450      \\ 
Cs  & 1,200         &  1,000     &   1,200        &   1,200    &  1,200       & - \\ 
De  & 1,900      &  1,200     &  2,000        &  1,200    & 1,200        & - \\  
Fr  &  1,400       &  1,200     &  1,200        & 1,050   & 1,165     & 700      \\  
Zh  & 2,550    & 1,200     &   1,200       & 1,165    & 2,550        & 1,100     \\  
Uk  & 1,000                & - & -   & 700   &  1,000        &  1,000     \\  \bottomrule[1pt]
\end{tabular}
}
\caption{The number of samples w.r.t different source-target language pairs. ``\emph{Src}'' and ``\emph{Tgt}'' denote the source and the target languages, respectively.}
\label{table:statistics1}
\end{table}

\section{Experimental Setup}

\begin{table}[t]
\centering
\resizebox{0.35\textwidth}{!}
{
  \begin{tabular}{lccc}
  \toprule[1pt]
  \multicolumn{1}{c}{LLM} & \multicolumn{1}{c}{Para.} & \multicolumn{1}{c}{Max Len.}   & \multicolumn{1}{c}{Flores}    \\ \midrule[1pt]
  \modi{gpt-4o-0816}              & \modi{-}                         & \modi{16K}          & \modi{\textbf{29.1}} \\
  gpt-4-1106              & -                         & 16K          & 27.7 \\
  gpt-3.5-turbo-1106      & -                         & 16K           & 22.0          \\
  LLaMa-2-13B-chat        & 13B                       & 4K         & 6.2           \\
  LLaMa-2-7B-chat         & 7B                        & 4K         & 5.2           \\
  \modi{LLaMa-3-8B-chat}         & \modi{8B}                        & \modi{8K}         & \modi{9.4}           \\
  Vicuna-13B-v1.5         & 13B                       & 4K           & 7.2           \\
  Vicuna-13B-v1.5-16k     & 13B                       & 16K            & 6.8           \\
  Vicuna-7B-v1.5          & 7B                        & 4K                & 6.1           \\
  Vicuna-7B-v1.5-16k      & 7B                        & 16K             & 5.9           \\
  Baichuan2-13B-Chat      & 13B                       & 4K               & 12.6          \\
  Baichuan2-7B-Chat       & 7B                        & 4K               & 11.0          \\
  Qwen-14B-Chat           & 14B                       & 8K          & 17.1          \\
  Qwen-7B-Chat            & 7B                        & 8K          & 13.2          \\
  \modi{Qwen2.5-14B-Chat}           & \modi{14B}                       & \modi{32K}          & \modi{19.2}          \\
  \modi{Qwen2.5-7B-Chat}            & \modi{7B}                        & \modi{32K}          & \modi{15.3}          \\
  Internlm2-chat-20B      & 20B                       & 32K         & 16.9          \\
  Internlm2-chat-7B       & 7B                        & 32K         & 15.2          \\ \bottomrule[1pt]
  \end{tabular}
}
\caption{Comparisons among LLMs used in experiments, including their parameters (Para.), maximum support length (Max Len.), and multi-lingual performance on Flores.}
\label{table:benchmarks}
\end{table}

\subsection{Evaluation LLMs}
We conduct experiments on the following three types of backbones, \emph{i.e.}, traditional models, closed- and open-source LLMs.

\vspace{0.2ex}
\noindent \textbf{Traditional multi-lingual models.}
(1) mBART-50~\cite{tang-etal-2021-multilingual} is a pre-trained multi-lingual model with transformer encoder-decoder architecture~\cite{vaswani2017attention} and 610M parameters.
(2) PISCES (610M)~\cite{wang-etal-2023-towards-unifying} is an M2MS pre-trained model that extends mBART-50 by further pre-training.

\vspace{0.2ex}
\noindent \textbf{Closed-source LLMs.}
(1) GPT-3.5-turbo (ChatGPT)~\cite{ChatGPT} is created by fine-tuning a GPT-3.5 series model via reinforcement learning from human feedback (RLHF). We use \emph{gpt-3.5-turbo-1106} in our experiments.
(2) GPT-4~\cite{OpenAI2023GPT4TR} is another advanced LLM that exhibits human-level performance on various benchmark datasets. We use \emph{gpt-4-1106} in our experiments.
\modi{(3) GPT-4o~\cite{hurst2024gpt} is an autoregressive omni model, which accepts multi-modal inputs and can generate multi-modal outputs.
The model shows superiority performance in various NLP benchmarks. We use \emph{gpt-4o-2024-08-16} in our experiments.}

\vspace{0.2ex}
\noindent \textbf{Open-source LLMs.}
(1) LLaMa~\cite{touvron2023llama2,dubey2024llama} is a LLM family, which also shows remarkable performance as a general task solver.
We use \emph{LLaMa-2-7B-chat}, \emph{LLaMa-2-13B-chat} and \modi{\emph{LLaMa-3-8B-chat}} in experiments.
(2) Vicuna~\cite{vicuna2023} is created by fine-tuning LLaMa-series models on user-shared conversations collected from ShareGPT.
Vicuna also provides 16k versions to support long text. We use \emph{Vicuna-7B-v1.5}, \emph{Vicuna-7B-v1.5-16k}, \emph{Vicuna-13B-v1.5} and \emph{Vicuna-13B-v1.5-16k} in experiments.
(3) BaiChuan-2~\cite{yang2023baichuan} is also trained with instruction-tuning and RLHF. This model shows its superior multi-lingual abilities in downstream tasks. We use \emph{Baichuan2-7B-Chat} and \emph{Baichuan2-13B-Chat} in experiments.
(4) Qwen~\cite{bai2023qwen,qwen2} is a LLM family, which shows great performance as a general task solver. We use \emph{Qwen-7B-Chat}, \emph{Qwen-14B-Chat}, \modi{\emph{Qwen2.5-7B-Chat} and \emph{Qwen2.5-14B-Chat}} in experiments.
(5) InternLM~\cite{team2023internlm} is a multi-lingual LLM pre-trained on multi-lingual corpora. We use \emph{Internlm2-chat-7B} and \emph{Internlm2-chat-20B} in experiments.

To provide a deeper understanding of the above LLMs, Table~\ref{table:benchmarks} compares their multi-lingual performance on Flores-101~\cite{goyal2021flores}, parameters and maximum support lengths.
To evaluate LLMs on M2MS, there are two settings should be considered:
(1) In the zero-shot prompting setting, LLMs directly perform M2MS based on a carefully designed prompt with task-specific instruction and in-context examples (Appenidx~\ref{sec:appendix_m2ms_prompt}).
(2) In the instruction-tuning setting, the training samples will be used for tuning open-source LLMs.
The above prompt is also used to formulate the M2MS sample into an instruction-response format.

\subsection{Evaluation Metrics}
We adopt ROUGE-1 (\textbf{R1}), ROUGE-2 (\textbf{R2}), ROUGE-L (\textbf{RL})~\cite{lin-2004-rouge} and BERTScore (\textbf{BS})~\cite{Zhang2020BERTScoreET}. The ROUGE scores measure the lexical overlap between the generated summaries and the references. BERTScore measures the similarity between them from a semantics perspective.
\modi{Besides, following~\citet{wang-etal-2023-chatgpt,liu-etal-2024-sumsurvey}, we prompt GPT-4o to score the generated summaries in terms of conciseness (\textbf{Con.}), coherence (\textbf{Coh.}), and relevance (\textbf{Rel.}) on a 5-point scale (more details are given in Appendix~\ref{sec:appendix_id_em}).}

\begin{table*}[t]
\centering
\resizebox{1.0\textwidth}{!}
{
 \begin{tabular}{lcccccc}
\toprule[1pt]
\multicolumn{1}{c}{\multirow{2}{*}{LLM}} & \textbf{Overall}                             & News                             & Encyc.                           & Dialogue                          & Guide                            & Tech.                            \\
\multicolumn{1}{c}{}                     & (R1 / RL / BS)         & (R1 / RL / BS)         & (R1 / RL / BS)         & (R1 / RL / BS)          & (R1 / RL / BS)         & (R1 / RL / BS)         \\ \midrule[1pt]
\multicolumn{7}{c}{\textbf{Setting 1: Zero-Shot LLMs}}                                                                                                                                                                                                                 \\ \midrule[1pt]
\modi{GPT-4o}    & \textbf{26.0} / \textbf{16.6} / \textbf{66.7} & \textbf{19.8} / \textbf{12.9} / \textbf{66.8} & 27.9 / \textbf{16.3} / \textbf{66.0} & \textbf{29.5} / \textbf{22.1} / \textbf{70.4} & \textbf{25.1} / \textbf{16.1} / \textbf{69.0} & \textbf{34.2} / 19.1 / 69.1  \\
GPT-4    & 25.7 / 16.4 / 66.4 & 19.5 / 12.5 / 65.9 & 26.9 / 14.5 / 64.8 & 28.9 / 21.6 / 70.0 & 24.0 / 15.5 / 68.4 & 33.8 / 18.8 / 68.9  \\
GPT-3.5-turbo     & 25.2 / 16.1 / \textbf{66.7} & 19.3 / 12.4 / 66.5 & \textbf{28.1} / 16.1 / 65.8 & 24.0 / 18.5 / 66.4 & 22.4 / 14.6 / 67.9 & 33.6 / \textbf{19.3} / \textbf{69.2}  \\
LLaMa-2-13B   & 21.5 / 13.0 / 64.1 & 17.9 / 11.8 / 65.1 & 25.5 / 14.6 / 63.8 & 19.3 / 13.8 / 64.0 & 18.4 / 11.3 / 64.3 & 30.7 / 17.2 / 64.5  \\
LLaMa-2-7B   & 18.2 / 10.8 / 63.3 & 14.2 / 09.0 / 64.0 & 22.7 / 12.7 / 62.4 & 17.4 / 12.7 / 62.6 & 15.0 / 09.3 / 63.6 & 23.6 / 13.8 / 62.5  \\
\modi{LLaMa-3-8B}   & 19.5 / 12.4 / 63.5 & 14.9 / 09.5 / 64.6 & 23.3 / 13.3 / 62.8 & 18.0 / 13.1 / 62.8 & 15.7 / 10.0 / 64.2 & 24.1 / 14.4 / 63.0  \\
Vicuna-13B      & 22.4 / 13.4 / 65.5 & 18.5 / 11.8 / 65.1 & 25.9 / 14.9 / 64.9 & 22.5 / 16.5 / 65.9 & 18.8 / 11.9 / 64.8 & 32.5 / 18.0 / 68.6  \\
Vicuna-13B-16k   & 22.9 / 13.9 / 66.0 & 19.0 / 11.9 / 65.3 & 27.2 / 15.5 / 65.3 & 22.6 / 17.1 / 66.1 & 20.3 / 12.9 / 65.9 & 33.0 / 19.2 / 69.1  \\
Vicuna-7B   & 22.3 / 13.7 / 65.0 & 17.8 / 11.6 / 65.5 & 26.0 / 15.4 / 64.9 & 22.1 / 16.2 / 67.1 & 18.5 / 11.3 / 65.3 & 31.1 / 17.5 / 67.4  \\
Vicuna-7B-16k    & 22.8 / 14.1 / 65.3 & 18.3 / 12.0 / 65.8 & 27.0 / 15.1 / 65.1 & 21.6 / 16.2 / 66.1 & 19.2 / 11.7 / 64.5 & 31.9 / 18.2 / 66.7  \\
Baichuan2-13B    & 20.5 / 12.8 / 65.0 & 15.9 / 10.2 / 64.9 & 24.4 / 13.7 / 64.1 & 19.8 / 15.3 / 64.9 & 18.1 / 11.1 / 65.0 & 30.0 / 16.8 / 66.2  \\
Baichuan2-7B    & 20.8 / 13.2 / 65.1 & 16.5 / 10.5 / 65.3 & 24.6 / 14.1 / 64.2 & 21.4 / 16.2 / 66.0 & 17.8 / 11.1 / 64.9 & 30.1 / 16.1 / 64.8   \\
Qwen-14B  & 21.6 / 13.0 / 65.2 & 17.9 / 11.5 / 65.6 & 25.3 / 14.3 / 64.7 & 21.8 / 16.3 / 66.5 & 18.1 / 11.1 / 64.7 & 32.0 / 17.8 / 64.8  \\
Qwen-7B     & 21.8 / 13.1 / 64.9 & 18.3 / 11.5 / 66.0 & 25.9 / 15.1 / 65.1 & 21.3 / 15.9 / 66.4 & 17.8 / 10.9 / 65.2 & 30.8 / 17.8 / 65.6 \\
\modi{Qwen2.5-14B}  & 22.1 / 13.1 / 65.4 & 18.4 / 11.7 / 65.8 & 25.8 / 14.8 / 65.2 & 22.0 / 16.6 / 66.8 & 18.5 / 11.6 / 64.9 & 32.6 / 18.1 / 65.2  \\
\modi{Qwen2.5-7B}     & 21.9 / 13.3 / 65.1 & 18.6 / 11.8 / 66.5 & 26.5 / 15.4 / 65.4 & 21.9 / 16.1 / 66.6 & 18.1 / 10.9 / 65.6 & 30.9 / 18.0 / 65.7 \\
Internlm2-20B  & 19.2 / 12.0 / 62.9 & 14.9 / 09.6 / 62.7 & 24.0 / 13.9 / 63.8 & 11.6 / 08.8 / 59.1 & 16.2 / 10.1 / 63.0 & 30.6 / 17.5 / 66.6  \\
Internlm2-7B   & 18.5 / 11.6 / 62.2 & 14.3 / 09.5 / 62.6 & 23.9 / 13.3 / 63.2 & 11.6 / 09.1 / 58.4 & 16.2 / 09.9 / 62.4 & 29.7 / 17.7 / 64.1  \\ \midrule[1pt]
\multicolumn{7}{c}{\textbf{Setting 2: Fine-Tuned Traditional Multi-Lingual Language Models}}                                                                                                                                                                                          \\ \midrule[1pt]
mBART-50   & 27.4 / 19.9 / 67.8 & \textbf{27.2} / \textbf{20.1} / 67.8 & 26.6 / 20.1 / 65.3 & 32.9 / 24.9 / \textbf{71.0} & 25.8 / 19.5 / 68.1 & 23.2 / 16.7 / 65.4   \\
PISCES   & \textbf{30.8} / \textbf{22.8} / \textbf{68.6} & \textbf{27.2} / 19.8 / \textbf{68.7} & \textbf{28.2} / \textbf{20.9} / \textbf{66.0} & \textbf{34.1} / \textbf{26.8} / 70.9 & \textbf{36.3} / \textbf{28.8} / \textbf{71.9} & \textbf{24.3} / \textbf{17.4} / \textbf{65.7}   \\ \midrule[1pt]
\multicolumn{7}{c}{\textbf{Setting 3: Instruction-Tuned LLMs}}                                                                                                                                                                                          \\ \midrule[1pt]
LLaMa-2-13B             & 37.7 / 29.4 / \textbf{74.4} & \textbf{37.1} / 27.2 / 74.2 & 40.2 / 32.2 / 74.2 & 40.3 / 32.4 / 75.4 & 33.0 / 26.6 / 73.4 & 38.2 / 26.2 / 73.4  \\
LLaMa-2-7B    & 35.5 / 27.0 / 73.0 & 34.9 / 26.5 / 73.2 & 37.6 / 29.2 / 73.2 & 37.9 / 30.8 / 75.0 & 31.8 / 25.9 / 72.6 & 37.8 / 25.7 / 72.7   \\
\modi{LLaMa-3-8B}    & 36.2 / 27.5 / 73.4 & 35.4 / 26.9 / 73.4 & 37.9 / 29.8 / 73.9 & 38.6 / 31.5 / 75.7 & 32.5 / 26.4 / 73.2 & 38.5 / 26.2 / 73.3   \\
Vicuna-13B   & 37.3 / 28.7 / 73.9 & 36.3 / 28.0 / 73.6 & 39.8 / 32.3 / \textbf{74.4} & 40.4 / 32.3 / 75.9 & 34.2 / 28.1 / 73.5 & 38.4 / \textbf{26.6} / 73.7   \\
Vicuna-13B-16k    & \textbf{38.0} / \textbf{30.2} / 74.1 & 36.9 / \textbf{28.6} / \textbf{74.7} & \textbf{40.4} / \textbf{32.9} / 74.2 & \textbf{41.2} / 33.6 / 75.9 & \textbf{34.5} / 28.5 / 73.9 & 38.3 / 26.4 / 73.5   \\
Vicuna-7B       & 35.6 / 28.0 / 73.1 & 34.5 / 26.2 / 73.8 & 38.3 / 29.1 / 73.3 & 38.9 / 32.3 / 75.4 & 31.2 / 25.7 / 72.9 & 36.8 / 24.5 / 72.4  \\
Vicuna-7B-16k    & 36.2 / 28.6 / 73.7 & 35.5 / 27.7 / 73.4 & 38.7 / 30.3 / 74.2 & 38.9 / 31.8 / 75.3 & 32.3 / 25.9 / 72.2 & 37.7 / 26.5 / 73.3  \\
Baichuan2-13B    & 36.1 / 28.0 / \textbf{74.4} & 35.9 / 26.4 / 73.1 & 38.0 / 29.8 / 73.2 & 40.7 / 34.1 / 75.9 & 33.5 / 25.7 / 73.7 & 39.2 / 25.3 / 73.8   \\
Baichuan2-7B      & 35.0 / 27.4 / 73.5 & 35.4 / 26.6 / 73.0 & 37.3 / 29.4 / 73.3 & 38.8 / 31.5 / 74.9 & 31.8 / 23.9 / 72.6 & 38.1 / 25.9 / 73.7   \\
Qwen-14B    & 37.1 / 28.4 / 74.2 & 36.0 / 26.8 / 73.2 & 38.4 / 30.8 / 73.4 & 40.2 / 33.4 / 75.5 & 34.4 / 28.5 / 74.4 & 39.1 / 26.4 / 74.4   \\
Qwen-7B     & 34.8 / 26.8 / 73.2 & 33.5 / 25.0 / 72.5 & 36.1 / 27.6 / 73.0 & 38.9 / 31.5 / 75.2 & 33.1 / 25.7 / 73.2 & 37.0 / 24.3 / 73.0   \\
\modi{Qwen2.5-14B}    & 37.8 / 29.3 / 74.3 & 36.7 / 28.0 / 74.0 & 39.4 / 31.6 / 73.9 & 40.8 / 33.6 / 75.8 & 34.4 / \textbf{28.8} / \textbf{74.7} & \textbf{39.4} / \textbf{26.6} / \textbf{74.8}   \\
\modi{Qwen2.5-7B}     & 35.2 / 27.3 / 73.7 & 34.2 / 25.6 / 72.9 & 36.6 / 28.2 / 73.7 & 39.5 / 32.3 / 76.0 & 33.4 / 26.5 / 73.6 & 37.7 / 24.9 / 73.6   \\
Internlm2-20B   & 36.7 / 28.2 / 73.7 & 35.0 / 25.6 / 73.0 & 38.0 / 29.0 / 72.8 & 41.1 / \textbf{34.2} / \textbf{76.2} & \textbf{34.5} / 27.9 / 74.3 & 39.3 / 25.7 / 73.4  \\
Internlm2-7B   & 35.7 / 27.2 / 73.5 & 34.1 / 25.6 / 72.6 & 36.2 / 28.4 / 72.8 & 40.7 / 33.5 / 75.9 & 33.3 / 27.0 / 73.3 & 37.8 / 25.6 / 73.0  \\ \bottomrule[1pt]
\end{tabular}
}
\caption{Experimental results of the overall performance and fine-grained results in each domain. The \textbf{bold} denotes the best performance under each setting. Encyc.: Encyclopedia; Tech.: Technology.}
\label{table:main_result}
\end{table*}

\subsection{Implementation Details}

For all LLMs, including open- and closed-source LLMs, we use the sampling decoding strategy, and follow \citet{liu2023alignbench} to set the temperature to 0.1. Besides, the maximum generation length is set to 400 tokens (the length of more than 97.8\% of summaries in used data is less than 400 tokens).
Since different LLMs have different maximum support lengths, we truncate the input source-language document to ensure the input length is within 3600 tokens to ensure fair comparison.
Please refer to Appendix~\ref{sec:appendix_id_llms} for more details of utilized model checkpoints, instruction-tuning LLMs, fine-tuning traditional models, and training hours.

\section{Results and Analyses}
\label{sec:result}
Table~\ref{table:main_result} shows the experimental results in terms of R1, RL and BS. For the full results including R2, please refer to Appendix~\ref{appendix:full_results}.
We analyze the results from the following aspects:

\noindent \textbf{Comparing traditional models with LLMs.}
When comparing traditional multi-lingual language models (mBART-50 and PISCES) with zero-shot LLMs, the fine-tuned traditional models slightly outperform the best zero-shot LLM (\emph{i.e.}, GPT-4o).
For example, GPT-4o achieves 26.0 R1 and 66.7 BS scores in overall performance, while the counterparts of mBART-50 are 27.4 and 67.8. PISCES outperforms mBART-50, and achieves 30.8 R1 and 68.6 BS scores.
This finding is consistent with previous exploration of other summarization tasks~\cite{wang-etal-2023-towards-unifying,qin2023chatgpt}.
The fine-tuned models learn the mapping from documents to summaries based on the whole training samples. In contrast, zero-shot LLMs only know a few in-context examples when performing M2MS.
Without parameter updating, zero-shot LLMs could achieve competitive results with fine-tuned traditional models, showing their powerful instruction-following and in-context learning abilities.

Comparing traditional models with instruction-tuned LLMs, we find that instruction-tuned LLMs generally outperform the best traditional language model (\emph{i.e.}, PISCES) by a large margin. For example, the instruction-tuned Vicuna-13B-16k outperforms PISCES by 7.2 R1, 7.4 RL and 5.5 BS scores. We analyze this phenomenon from the following aspects:
(1) The LLMs involve more parameters than the traditional models, thus having a more powerful ability to fit the tasks-specific data with small-scale instruction samples. As shown in previous work~\cite{ladhak-etal-2020-wikilingua,perez-beltrachini-lapata-2021-models,bhattacharjee-etal-2023-crosssum}, traditional models need a large number of multi-lingual summarization samples to learn how to generate a target-language summaries for the given source-language documents. Even in a single domain, traditional models generally need more than 100K samples during their training stage~\cite{liang-etal-2022-variational}. However, our study only uses 19.5K training samples from five domains, resulting in a great challenge to traditional models.
(2) Another important aspect is the model's maximum support length. Many documents in M2MS samples contain more than 2K tokens which is larger than the maximum support length of traditional models (c.f., Appendix~\ref{sec:appendix_data_statistics}).
For example, mBART-50 and PISCES only support input text within 1K tokens due to the limitation in their vanilla $\mathcal{O}(n^2)$ self-attention mechanism~\cite{vaswani2017attention}. In contrast, LLMs typically adopt RoPE~\cite{su2024roformer} that comes with valuable properties such as the flexibility of being expanded to any sequence length, and the capability of equipping the linear self-attention with relative position encoding. In this manner, LLMs could support long-document inputs and are more practical in real-world scenes.

\vspace{0.5ex}
\noindent \textbf{Comparisons among zero-shot LLMs.}
Among all zero-shot LLMs, GPT-4o achieves the best results in overall performance while GPT-4 achieves the second results in most cases.
Compared with other LLMs, GPT-4o shows its powerful ability to follow human instructions to perform MSMS and generally outperforms other LLMs.
Among open-source LLMs, Vicuna-13B-16k works best and reaches 22.9 R1, 13.9 RL and 66.0 BS scores in overall performance, verifying the effectiveness of instruction-tuning LLaMa-series LLMs with the ShareGPT's user conversations.
Comparing Vicuna-series LLMs with other open-source LLMs, we find that though the multi-lingual ability of Vicuna-series LLMs is less than others (as demonstrated in Table~\ref{table:benchmarks}), Vicuna-series LLMs typically outperform others in M2MS, showing its superior instruction-following and in-context learning abilities distilled from ShareGPT.
In addition, we also find that Vicuna-13B-16k and Vicuna-7B-16k outperform Vicuna-13B and Vicuna-7B, respectively. Though the input length is truncated to 4K tokens for all LLMs, the Vicuna-16k models scale the support length from 4K to 16K and achieve better behaviors when processing long documents.

\begin{figure}[t]
\centering
\subfigure{
  \includegraphics[width=0.8\linewidth]{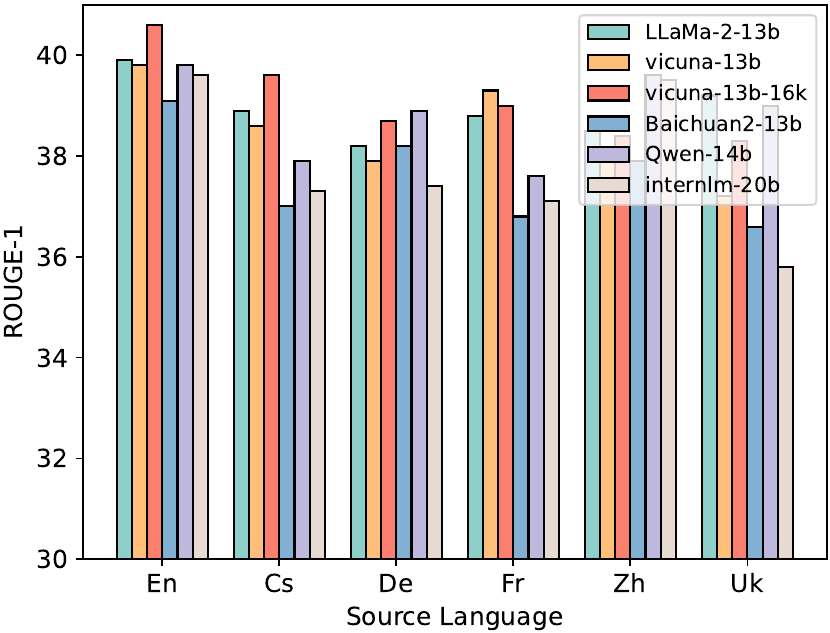}
}
\subfigure{
  \includegraphics[width=0.86\linewidth]{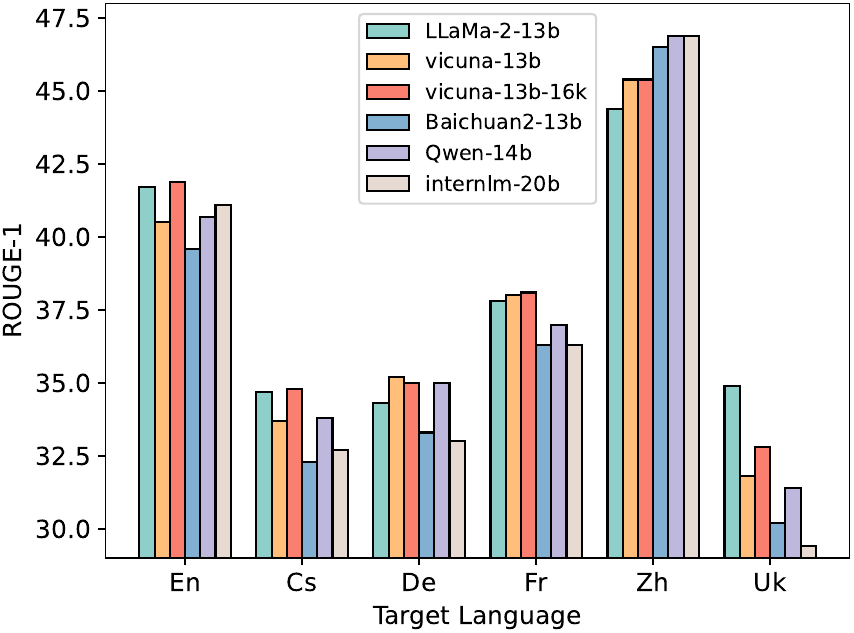}
}
\caption{Language-wise performance of tuned LLMs.}
\label{fig:lang_wise_performance}
\end{figure}

\vspace{0.5ex}
\noindent \textbf{Comparisons among instruction-tuned LLMs.}
Vicuna-13B-16k also performs best among all instruction-tuned LLMs, and achieves 38.0 R1, 30.2 RL and 74.1 BS scores in overall performance.
Qwen2.5-14B and LLaMa-2-13B achieve promising results following closely behind Vicuna-13B-16k.
Besides, all instruction-tuned LLMs significantly outperform the fine-tuned traditional models as well as zero-shot LLMs by a large margin.
This finding demonstrates that LLMs can improve their task-specific ability via instruction-tuning on small-scale task data.
In addition to domain-wise performance, we also show the instruction-tuned LLM performance when using a specific source or target language during evaluation. The results are provided in Figure~\ref{fig:lang_wise_performance}.
Different LLMs might be good at different languages. For example, Vicuna-13B-16k performs best when processing English and Czech documents, while Qwen-14b performs best in German and Chinese documents. As for target languages, the LLaMa and Vicuna LLMs are good at generating English, Czech, French and Ukrainian summaries. Baichuan2-13B, Qwen-14B and Internlm-20B do well in generating Chinese summaries.

\begin{table}[t]
\centering
\resizebox{0.48\textwidth}{!}
{
\begin{tabular}{lccclccc}
\toprule[1pt]
& \multicolumn{3}{c}{Zero-shot Results} &                & \multicolumn{3}{c}{Ins-tuned Results} \\
& Con. & Coh. & Rel. &                & Con. & Coh. & Rel. \\ \midrule[1pt]
\modi{GPT-4o}          & 3.30  & \textbf{4.66} & \textbf{4.83} &        &   &  &  \\
GPT-4          & 3.27  & 4.63 & 4.75 & mBART-50       & 3.82  & 3.42 & 4.02 \\
GPT-3.5-turbo  &  3.04     &  4.56    & 4.69     & PISCES         & 3.95      &  3.58    &  4.15    \\
LLaMa-2-13B    &  3.20     &  4.34    &  4.50    & LLaMa-2-13B    & 4.59      &  4.70    &   \textbf{4.72}   \\
LLaMa-2-7B     &  3.28     &  4.29    &  4.39    & LLaMa-2-7B     & 4.42      &   4.57   &  4.65    \\
\modi{LLaMa-3-8B}     &  3.32     &  4.31    &  4.48   & \modi{LLaMa-3-8B}     & 4.50      &   4.59   &  4.68    \\
Vicuna-13B     &  3.15     &  4.35    &  4.51    & Vicuna-13B     & \textbf{4.70}      &  4.68    &  4.70    \\
Vicuna-13B-16k &  3.10     &  4.39    & 4.52     & Vicuna-13B-16k & 4.63      &  \textbf{4.71}    &  4.66    \\
Vicuna-7B      &  3.06     &  4.27    & 4.45     & Vicuna-7B      & 4.47      &  4.55    &  4.62    \\
Vicuna-7B-16k  &  3.10     &  4.30    & 4.37     & Vicuna-7B-16k  & 4.53      &  4.58    &  4.60    \\
Baichuan2-13B  &  \textbf{3.37}     &  4.34    &  4.46    & Baichuan2-13B  & 4.51      &  4.65    &   4.69   \\
Baichuan2-7B   &  3.19     &  4.25    &  4.29    & Baichuan2-7B   & 4.40      &  4.51    &  4.53    \\
Qwen-14B       &  3.25     &  4.32    &  4.40    & Qwen-14B       & 4.55      &  4.64    &  4.60    \\
Qwen-7B        &  3.30     &  4.23    &  4.33    & Qwen-7B        & 4.39      &  4.51    &  4.45    \\
\modi{Qwen2.5-14B}       &  3.32     &  4.37    &  4.48    & \modi{Qwen2.5-14B}       & 4.63      &  4.69    &  4.65    \\
\modi{Qwen2.5-7B}        &  3.27     &  4.29    &  4.32    & \modi{Qwen2.5-7B}        & 4.42      &  4.56    &  4.50    \\
Internlm2-20B  &  3.17     &  4.29    &  4.37    & Internlm2-20B  & 4.51      &  4.60    &  4.67    \\
Internlm2-7B   &  3.04     &  4.18    &  4.28    & Internlm2-7B   & 4.25      &  4.47     & 4.52     \\ \bottomrule[1pt]
\end{tabular}
}
\caption{\modi{Overall performance in terms of GPT-4o evaluation (Ins-tuned: Instruction-tuned).}}
\label{table:gpt4oscore}
\end{table}

\vspace{0.5ex}
\noindent \textbf{LLM evaluation results.}
\modi{Table~\ref{table:gpt4oscore} shows the evaluation results using GPT-4o evaluation.
The instruction-tuned LLMs significantly outperform traditional models in all metrics.
For zero-shot LLMs, they tend to generate relatively lengthy summaries compared to traditional models or fine-tuned LLMs, resulting in low conciseness. This is because the alignment phase in LLMs emphasizes the usefulness of the models, making them provide detailed information.
After tuning, LLMs can study to generate short summaries, and significantly improve their conciseness scores.
In terms of coherence and relevance, instruction-tuning also brings improvements to LLMs.
Moreover, we find that some tuned LLMs (\emph{e.g.}, LLaMa-2-7B and Vicuna-7B) achieve lower coherence and relevance than zero-shot GPT-4, though they outperform GPT-4 in terms of ROUGE scores and BERTScore.}

\vspace{0.5ex}
\noindent \textbf{The effects of training scales.}
\modi{In our main experiments, we use 19.5K training samples to fine-tune LLMs and traditional models.
We further discuss the effects of the training scales on model performance, please refer to Appendix~\ref{appendix:effects_of_ts}.}

\begin{table}[t]
\centering
\resizebox{0.45\textwidth}{!}
{
\begin{tabular}{lclc}
\toprule[1pt]
\multicolumn{1}{c}{Model} & Rate & \multicolumn{1}{c}{Model} & Rate \\ \midrule[1pt]
LLaMa-2-13B    & 91.1$^\dagger$ / 98.7$^\ddagger$ & Baichuan2-13B & 98.7$^\dagger$ / 99.4$^\ddagger$ \\
LLaMa-2-7B     & 81.6$^\dagger$ / 98.7$^\ddagger$ & Baichuan2-7B  & 94.8$^\dagger$ / 99.4$^\ddagger$ \\
\modi{LLaMa-3-8B}     & 96.6$^\dagger$ / 98.8$^\ddagger$ & Qwen-14B      & 98.7$^\dagger$ / 99.3$^\ddagger$  \\
Vicuna-13B     & 96.6$^\dagger$ / 98.6$^\ddagger$ & Qwen-7B       & 98.7$^\dagger$ / 99.1$^\ddagger$ \\
Vicuna-13B-16k & 98.7$^\dagger$ / 98.8$^\ddagger$ & \modi{Qwen2.5-14B} & 99.2$^\dagger$ / 99.4$^\ddagger$ \\
Vicuna-7B      & 94.5$^\dagger$ / 99.2$^\ddagger$ & \modi{Qwen2.5-7B} & 98.5$^\dagger$ / 99.1$^\ddagger$ \\
Vicuna-7B-16k  & 97.5$^\dagger$ / 98.7$^\ddagger$ & Internlm2-20B & 97.0$^\dagger$ / 98.7$^\ddagger$ \\ 
  &  & Internlm2-7B  & 97.2$^\dagger$ / 98.5$^\ddagger$ \\ \bottomrule[1pt]
\end{tabular}
}
\caption{Correct language rate (\%) of the summaries ($^{\dagger}$ and $^{\ddagger}$ denote the results of zero-shot and instruction-tuned LLMs, respectively).}
\label{table:cr_score_res}
\end{table}

\section{Discussion}

\noindent \textbf{Are generated summaries in the correct language?}
\modi{As reported by \citet{wang-etal-2023-towards-unifying}, using traditional models as multi-lingual summarizers might generate in the wrong languages instead of the given target language.
We wonder whether LLM-generated summaries are in the correct language.
To this end, we use \emph{fastlangid}\footnote{\url{https://pypi.org/project/fastlangid/}} to detect the generated summaries, and calculate the rate of the generated summaries in the correct language, named correct language rate (\textbf{CR}).
As shown in Table~\ref{table:cr_score_res}, we find that most zero-shot LLMs could follow the language requirements in the prompt and generate summaries in the right target language with $\geq$ 95 CR score. This is because the LLMs are trained with multi-lingual corpora that include a large number of parallel sentences across different languages, making the model already learn the multi-lingual skills.
After tuning, LLMs generally improve their CR scores.
Thus, using LLMs to serve as the backbones of the M2MS systems has great potential in real-world applications.}

\begin{table}[t]
\centering
\resizebox{0.44\textwidth}{!}
{
\begin{tabular}{lcclcc}
\toprule[1pt]
\multicolumn{1}{c}{LLM} & Before & After & \multicolumn{1}{c}{LLM} & Before & After \\ \midrule[1pt]
LLaMa-2-13B             & 53.5   & 54.3  & Baichuan2-13B           & 54.5   & 54.1  \\
LLaMa-2-7B              & 47.2   & 47.3  & Baichuan2-7B            & 52.9   & 52.0  \\
Vicuna-13B              & 55.5   & 55.8  & Qwen-14B                & 66.0   & 65.9  \\
Vicuna-13B-16k          & 54.3   & 54.5  & Qwen-7B                 & 56.2   & 57.0  \\
Vicuna-7B               & 49.8   & 49.6  & Internlm2-20B           & 65.0   & 64.6  \\
Vicuna-7B-16k           & 48.0   & 48.4  & Internlm2-7B            & 59.1   & 58.9  \\ \bottomrule[1pt]
\end{tabular}
}
\caption{The LLMs' performance on MMLU. ``\emph{Before}'' and ``\emph{After}'' denote the results of LLMs before and after the M2MS instruction tuning, respectively.}
\label{table:mmlu_res}
\end{table}

\vspace{0.5ex}
\noindent \textbf{Does the instruction tuning on M2MS influence LLMs' general ability?}
\modi{As shown in Section~\ref{sec:result}, LLMs can improve their M2MS ability by instruction-tuning on the collected training samples. An important question arises naturally, \emph{i.e.}, \emph{does this task-specific improvement sacrifice the general task-solving ability of LLMs?}
To figure out this question, we further evaluate the LLMs before and after M2MS instruction tuning on the MMLU evaluation dataset~\cite{hendrycks2021measuring}.
The MMLU dataset covers 57 tasks including elementary mathematics, US history, computer science, law, etc, and is designed to evaluate models' world knowledge and problem-solving ability.
We follow previous LLM work~\cite{touvron2023llama2,yang2023baichuan,bai2023qwen}, and adopt the 5-shot evaluation strategy. The experimental results are provided in Table~\ref{table:mmlu_res}.
As we can see, the M2MS instruction tuning on LLMs does not sacrifice their general task-solving ability. Some instruction-tuned LLMs (\emph{i.e.}, LLaMa-2-13B and Qwen-7B) even outperform their original models.}

\vspace{0.5ex}
\noindent \textbf{Can large model generate factually consistent summaries?}
\modi{As revealed by~\citet{maynez-etal-2020-faithfulness,gao-etal-2023-evaluating}, the model-generated summaries might be inconsistent with the source documents.
We want to know if LLMs can generate factually consistent summaries.
To this end, we randomly select 100 samples from the testing set and conduct fine-grained human evaluation on the summaries generated by GPT-4, zero-shot \& tuned LLaMa-2-13B, and zero-shot \& tuned Vicuna-13B-16k.
Following~\citet{gao-etal-2023-evaluating}, for each sample, we employ human evaluators to annotate the following four types of factual errors (if has):
(1) hallucination error: a generated summary contains events not directly inferable from the given document.
(2) particulars error: the summary contains the right events but some details are inaccurate or mistaken.
(3) predicate error: the predicate in the summary is contradictory to the source document.
(4) entity error: the entity of an event in the summary is wrong.
More details about human evaluation are given in Appendix~\ref{sec:appendx_human_evaluation}.
Table~\ref{table:human_eval_res} reports the proportion of each error.
The summaries generated by GPT-4 have the lowest error proportion in terms of all errors.
Besides, among all types of factual errors, the hallucination error occurs more frequently, indicating it is a non-trivial issue when adapting LLMs as the summarizers.
Another finding is that instruction tuning on LLMs might intensify their factual issue especially the hallucination and the particulars errors. We conjecture this because the summaries in the training samples are written by humans, thus they might involve more information (like background information) beyond the given documents.
Such an information gap might encourage LLMs to generate hallucinations during tuning.
As reported by previous studies~\cite{wang-etal-2022-analyzing,gao-etal-2023-evaluating}, the ground truth references in summarization data also have the hallucination error.
Therefore, when building LLM M2MS summarizers in the real applications, this error should be carefully considered during instruction tuning.}

\begin{table}[t]
\centering
\resizebox{0.40\textwidth}{!}
{
\begin{tabular}{lcccc}
\toprule[1pt]
                         & Hallu. & Parti. & Predi. & Entity \\ \midrule[1pt]
GPT-4                    & 8      & 3      & 5      & 3      \\
Zero-shot LLaMa-2-13B    & 17     & 13     & 9     & 12     \\
Tuned LLaMa-2-13B        & 23     & 18     & 7     & 9     \\
Zero-shot Vicuna-13B-16k & 12     & 8     & 8     & 10     \\
Tuned Vicuna-13B-16k     & 17     & 16     & 10     & 6     \\ \bottomrule[1pt]
\end{tabular}
}
\caption{Fine-grained human evaluation results on factuality (Hallu.: hallucination; Parti.: particulars; Predi.: predicate).}
\label{table:human_eval_res}
\end{table}

\section{Conclusion}

In this paper, we explore how well off-the-shelf LLMs can deal with the many-to-many summarization (M2MS) task.
Considering the limited diversity and the single domain characteristics in each single dataset, we reorganize M2MS data based on eight existing multi-lingual summarization datasets, and the used data covers five domains and six languages.
Based on it, we conduct extensive experiments on various open- and closed-source LLMs.
Our results indicate that the zero-shot LLMs could achieve competitive results with fine-tuned traditional models.
Furthermore, through instruction tuning, open-source LLMs can significantly improve their M2MS ability, and not sacrifice their general capabilities.
\modi{However, as shown in our human evaluation, LLMs still face the factuality issue, and the instruction tuning might intensify this issue, which is worth noting in future research.}

\section*{Limitations}
While we evaluate the performance of LLMs on the many-to-many summarization task, there are some limitations worth noting:
(1) We only evaluate the lower threshold of these models' M2MS performance. Prompts are important to guide LLMs to perform specific tasks, and future work could explore better prompts to obtain better results.
(2) The used data in our empirical study only involves data from existing multi-lingual summarization datasets. Future work could extend it with more domains in the real scenes.

\section*{Ethical Considerations}
In this paper, we use multiple LLMs (\emph{e.g.}, GPT-4o, GPT-4, GPT-3.5-turbo and LLaMa-2) and traditional language models (\emph{e.g.}, mBART) as the M2MS models in experiments. During instruction tuning and fine-tuning, the adopted M2MS samples mainly come from previous datasets, \emph{i.e.}, CrossSum~\cite{bhattacharjee-etal-2023-crosssum}, XWikis~\cite{perez-beltrachini-lapata-2021-models}, XSAMSum~\cite{wang-etal-2022-clidsum}, XMediaSum~\cite{wang-etal-2022-clidsum}, DialogSumX~\cite{chen-etal-2023-revisiting}, WikiLingua~\cite{ladhak-etal-2020-wikilingua}, Perseus~\cite{zheng2023long} and Spektrum~\cite{fatima-strube-2021-novel}.
Therefore, the trained models might involve the same biases and toxic behaviors exhibited by these datasets.

\bibliography{custom}

\appendix

\section{Details of Intrinsic Metrics}
\label{sec:appendix_intrinsic_metrics}

Following \citet{grusky-etal-2018-newsroom,bommasani-cardie-2020-intrinsic}, we filter out low-quality summarization samples based on the following three intrinsic quality metrics:

(1) \emph{Coverage} evaluates the percentage of words in a summary that are part of an extractive fragment from the document~\cite{grusky-etal-2018-newsroom}:
\begin{equation}
\small
\text{Coverage}(D, S) = \frac{1}{|S|} \sum_{u \in F(D, S)} |u|
\end{equation}
where $D$ and $S$ denote a document and the corresponding summary, respectively. $F(D, S)$ is the set of all extractive fragments that appear in both $D$ and $S$. A smaller number of coverage indicates the summary is more abstractive.

(2) \emph{Redundancy} measures whether sentences in a summary are similar to each other~\cite{bommasani-cardie-2020-intrinsic}. Assuming a summary $S$ has $m$ sentences $S=\{s_1, s_2, ..., s_m\}$, redundancy is formally calculated as follows: 
\begin{equation}
\small
\text{Redundancy}(S) = \frac{1}{{m \choose 2}} \sum_{1 \leq i \leq m-1} \sum_{i+1 \leq j \leq m} \text{RL} (m_i, m_j)
\end{equation}
where $\text{RL}(\cdot)$ denotes the ROUGE-L score~\cite{lin-2004-rouge}. Generally, the lower the redundancy, the higher the sample quality.

(3) \emph{Coherence} measures the semantic coherence of a summary $S=\{s_1, ..., s_m\}$ by predicting the probability of each successive sentence conditioned on the previous one using a language model~\cite{bommasani-cardie-2020-intrinsic}:
\begin{equation}
\small
\text{Coherence}(S) = \frac{1}{m-1} \sum^{m}_{i=2} P_\theta (s_i | s_{i-1})
\end{equation}
where $P_\theta$ denotes the predicted probability of a language model. Here, we adopt mBERT~\cite{devlin-etal-2019-bert} as $P_\theta$. Generally, a high coherence score indicates the high quality of the sample.

For each intrinsic metric, we set a threshold to filter low-quality samples. Specifically, inspired by~\citet{bommasani-cardie-2020-intrinsic}, for each sample, its coverage should be less than $\alpha_\text{cov} = 0.9$, the redundancy should be less than $\alpha_\text{red} = 0.2$, and the coherence needs to be more than $\alpha_\text{coh} = 0.9$. Otherwise, the sample will be filtered out.

\section{\modi{Data Contamination}}
\label{sec:appendix_data_contamination}

\modi{Data contamination is a potential major issue in measuring LLMs’ performance on downstream tasks, where the testing data might be in the pertaining corpora of LLMs~\cite{xu2024benchmark}.
\citet{golchin2024time} propose an effective method using BLEURT \& ROUGE-L metrics to measure instance-level contamination, \emph{i.e.}, identifying if an instance (usually a sentence or document) is contaminated for a given LLM.
The method can be used in both open-source and closed-source LLMs.}

\modi{When selecting the testing samples in our empirical study, after filtering samples via intrinsic metrics (Appendix~\ref{sec:appendix_intrinsic_metrics}), we calculate the instance-level contamination for each sample w.r.t GPT-4o, Vicuna-7B, Baichuan2-7B, Qwen2.5-7B-Chat, LLaMa-3-8B-chat and Internlm2-7B.\footnote{\modi{For the same series of LLMs, we select one to measure the contamination.}}
The uncontaminated samples will be randomly selected to form the testing set.
Some directions in the domain-specific datasets might contain only a few hundred samples (\emph{e.g.}, CrossSum only contains about 300 Fr$\Rightarrow$Zh samples), thus we cannot ensure data contamination and testing scale simultaneously in these directions.
Therefore, an extremely small number of samples labeled as ``contaminated'' will also be included in our testing set, and they account for less than 1\% of the whole testing set.
}

\begin{table*}[t]
\centering
\resizebox{0.98\textwidth}{!}
{
\begin{tabular}{ccccccc}
\toprule[1pt]
 \diagbox[dir=NW]{Src}{Tgt}  & En                           & Cs                 & De                   & Fr               & Zh                       & Uk                 \\ \midrule[1pt]
\multirow{2}{*}{En}  & 1400 / 1250 / 1250             & 500 / 350 / 350      & 1300 / 1050 / 1050     & 500 / 425 / 425  & 1150 / 950 / 950         & 700 / 375 / 375      \\
                     & (CR, XW, XS, XM, DI, WI, SP) & (XW, WI)           & (XW, XS, XM, WI, SP) & (CR, XW, DI, WI) & (CR, XW, XS, XM, DI, WI) & (CR, DI)           \\ \midrule
\multirow{2}{*}{Cs}  & 500 / 350 / 350                & 300 / 350 / 350      & 500 / 350 / 350        & 500 / 350 / 350    & 500 / 350 / 350            & \multirow{2}{*}{-} \\
                     & (XW, WI)                     & (XW, WI)           & (XW, WI)             & (XW, WI)         & (XW, WI)                 &                    \\ \midrule
\multirow{2}{*}{De}  & 800 / 550 / 550                & 500 / 350 / 350      & 900 / 550 / 550        & 500 / 350 / 350    & 500 / 350 / 350            & \multirow{2}{*}{-} \\
                     & (XW, WI, SP)                 & (XW, WI)           & (XW, WI, SP)         & (XW, WI)         & (XW, WI)                 &                    \\ \midrule
\multirow{2}{*}{Fr}  & 650 / 375 / 375                & 500 / 350 / 350      & 500 / 350 / 350        & 300 / 375 / 375    & 565 / 300 / 300            & 500 / 100 / 100      \\
                     & (CR, XW, WI)                 & (XW, WI)           & (XW, WI)             & (CR, XW, WI)     & (CR, XW, WI)             & (CR)               \\ \midrule
\multirow{2}{*}{Zh}  & 900 / 825 / 825              & 500 / 350 / 350      & 500 / 350 / 350        & 565 / 300 / 300    & 900 / 825 / 825          & 500 / 300 / 300      \\
                     & (CR, XW, WI, PE)             & (XW, WI)           & (XW, WI)             & (CR,  XW, WI)    & (CR, XW, WI, PE)         & (CR)               \\ \midrule
\multirow{2}{*}{Uk}  & 400 / 300 / 300                & \multirow{2}{*}{-} & \multirow{2}{*}{-}   & 400 / 150 / 150    & 400 / 300 / 300            & 400 / 300 / 300      \\
                     & (CR)                         &                    &                      & (CR)             & (CR)                     & (CR)               \\ \bottomrule[1pt]
\end{tabular}
}
\caption{The number of training/validation/testing samples and the data sources w.r.t different source-target language pairs. ``\emph{Src}'' and ``\emph{Tgt}'' denote the source and the target languages, respectively. (CR: CrossSum; XW: XWikis; XS: XSAMSum; XM: XMediaSum; DI: DialogSumX; WI: WikiLingua; PE: Perseus; SP: Spektrum)}
\label{appendix_table:statistics}
\end{table*}

\section{Data Statistics}
\label{sec:appendix_data_statistics}

\noindent \textbf{Language and Source Distribution.} For a specific language pair, the number of samples in each subset (training, validation, and testing sets) and the corresponding data sources are provided in Table~\ref{appendix_table:statistics}.

\begin{table}[t]
\centering
\resizebox{0.48\textwidth}{!}
{
\begin{tabular}{llrrrrrr}
\toprule[1pt]
\multicolumn{1}{l}{}        &      & \multicolumn{1}{c}{En$\rightarrow$X} & \multicolumn{1}{c}{CS$\rightarrow$X} & \multicolumn{1}{c}{De$\rightarrow$X} & \multicolumn{1}{c}{Fr$\rightarrow$X} & \multicolumn{1}{c}{Zh$\rightarrow$X} & \multicolumn{1}{c}{Uk$\rightarrow$X} \\ \midrule[1pt]
\multirow{2}{*}{Training}   & Doc. & 869.05                                 & 2418.82                                & 2111.35                                & 1385.84                                & 2534.56                                & 1693.38                                \\
                            & Sum. & 65.64                                  & 124.67                                 & 196.98                                 & 84.11                                  & 150.68                                 & 57.97                                  \\
\multirow{2}{*}{Validation} & Doc. & 897.45                                 & 2364.50                                & 2087.41                                & 1345.95                                & 2164.52                                & 1477.90                                \\
                            & Sum. & 62.34                                  & 130.32                                 & 177.85                                 & 88.44                                  & 141.26                                 & 57.21                                  \\
\multirow{2}{*}{Testing}    & Doc. & 791.15                                 & 2361.89                                & 2074.92                                & 1280.33                                & 2269.90                                & 1346.57                                \\
                            & Sum. & 52.32                                  & 125.68                                 & 198.89                                 & 86.41                                  & 134.35                                 & 52.02                                  \\ \midrule[1pt]
\multicolumn{1}{l}{}        &      & \multicolumn{1}{c}{X$\rightarrow$En} & \multicolumn{1}{c}{X$\rightarrow$Cs} & \multicolumn{1}{c}{X$\rightarrow$De} & \multicolumn{1}{c}{X$\rightarrow$Fr} & \multicolumn{1}{c}{X$\rightarrow$Zh} & \multicolumn{1}{c}{X$\rightarrow$Uk} \\ \midrule[1pt]
\multirow{2}{*}{Training}   & Doc. & 2105.74                                & 2204.62                                & 2080.01                                & 1627.83                                & 2102.88                                & 1050.10                                \\
                            & Sum. & 115.40                                  & 159.83                                 & 141.19                                 & 58.17                                  & 140.67                                 & 89.05                                  \\
\multirow{2}{*}{Validation} & Doc.  & 1731.99 & 2121.00 & 2307.03 & 1209.37 & 1821.54 & 982.90                      \\
    & Sum. & 98.52 & 167.79 & 124.08 & 57.89 & 129.91 & 89.89 \\
\multirow{2}{*}{Testing}    & Doc. & 1769.09 & 1807.71 & 2092.57 & 1502.02 & 1883.59 & 799.59        \\
   & Sum.  & 96.41 & 163.11 & 142.47 & 59.25 & 119.95 & 79.27   \\ \bottomrule[1pt]
\end{tabular}
}
\caption{The token-level average length of documents (Doc.) and summaries (Sum.) in the data w.r.t different source and target languages. En$\rightarrow$X/X$\rightarrow$En indicates all samples whose documents/summaries are in English.}
\label{table:statistics2}
\end{table}

\begin{table}[t]
\centering
\resizebox{0.48\textwidth}{!}
{
\begin{tabular}{llrrrrr}
\toprule[1pt]
&         & \multicolumn{1}{c}{News} & \multicolumn{1}{c}{Encyc.} & \multicolumn{1}{c}{Dialogue} & \multicolumn{1}{c}{Guide} & \multicolumn{1}{c}{Tech.} \\ \midrule[1pt]
\multirow{3}{*}{Training}   & Num. & 4680                     & 4650                             & 2550                         & 4650                      & 3000                           \\
    & Doc.    & 1350.21                  & 2020.66                          & 733.24                       & 772.6                     & 2857.5                         \\
    & Sum.    & 63.36                    & 121.46                           & 42.04                        & 69.96                     & 235.14                         \\ \midrule
\multirow{3}{*}{Validation} & Num. & 2300                      & 3425                              & 2600                         & 3425                      & 2400                            \\
    & Doc.     & 1101.92 & 1936.80 & 728.72 & 767.20 & 2687.42                     \\
    & Sum.    & 57.18 & 122.57 & 40.32 & 67.10 & 221.87                     \\ \midrule
\multirow{3}{*}{Testing}    & Num. & 2300                      & 3425                              & 2600                         & 3425                      & 2400    \\
    & Doc.    & 1055.44 & 1994.10 & 749.52 & 732.96 & 2882.51                     \\
    & Sum.    & 54.72 & 123.86 & 36.18 & 68.36 & 245.10                        \\ \bottomrule[1pt]
\end{tabular}
}
\caption{The token-level average length of documents (Doc.) and summaries (Sum.) in the data w.r.t different domains. ``\emph{Num.}'' indicates the number of samples in each domain. Encyc.: Encyclopedia; Tech.: Technology}
\label{table:statistics3}
\end{table}

\begin{figure}[t]
\centerline{\includegraphics[width=0.40\textwidth]{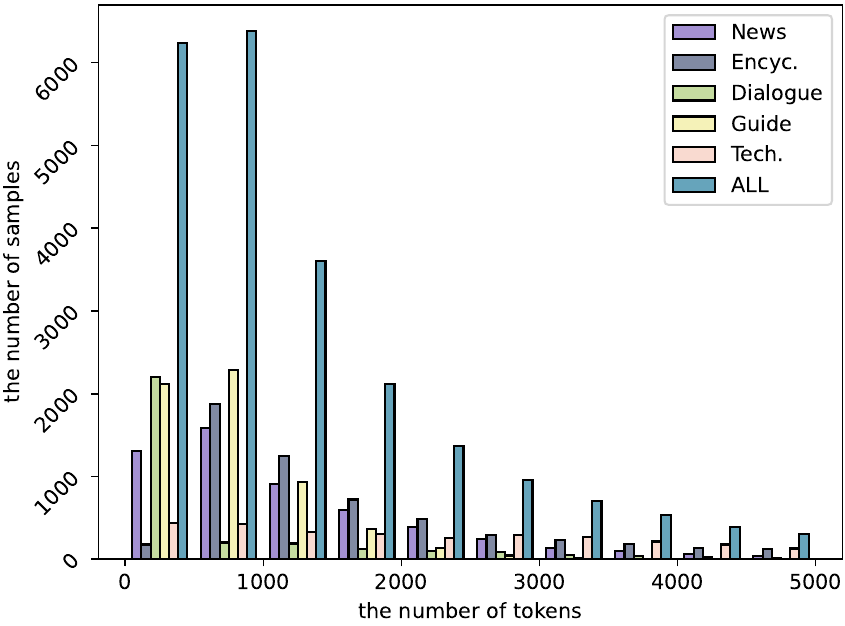}}
\caption{Length distributions of M2MS samples w.r.t different domains.}
\label{fig:length_distributions}
\end{figure}

\vspace{0.5ex}
\noindent \textbf{Length and Domain Distributions.} To calculate the length of documents and summaries across different languages, we use tiktoken\footnote{\url{https://github.com/openai/tiktoken}} to tokenize the documents and summaries, and calculate their token-level length.
As shown in Table~\ref{table:statistics2}, the average length of source documents typically reaches thousands of tokens, while the counterpart of target summaries is within 200 tokens.
From the perspective of domains, Table~\ref{table:statistics3} shows the average length of documents as well as summaries w.r.t different domains. As we can see, the average document length in the encyclopedia and technology domains is generally more than that in other domains.
The average length of dialogue and guide documents is less than 800 tokens, making them the shortest document length among all domains. 
To provide a deeper understanding of the used data in our empirical study, Figure~\ref{fig:length_distributions} shows the length distributions of different domains.

\section{M2MS Prompt}
\label{sec:appendix_m2ms_prompt}

\begin{figure}[t]
\centerline{\includegraphics[width=0.45\textwidth]{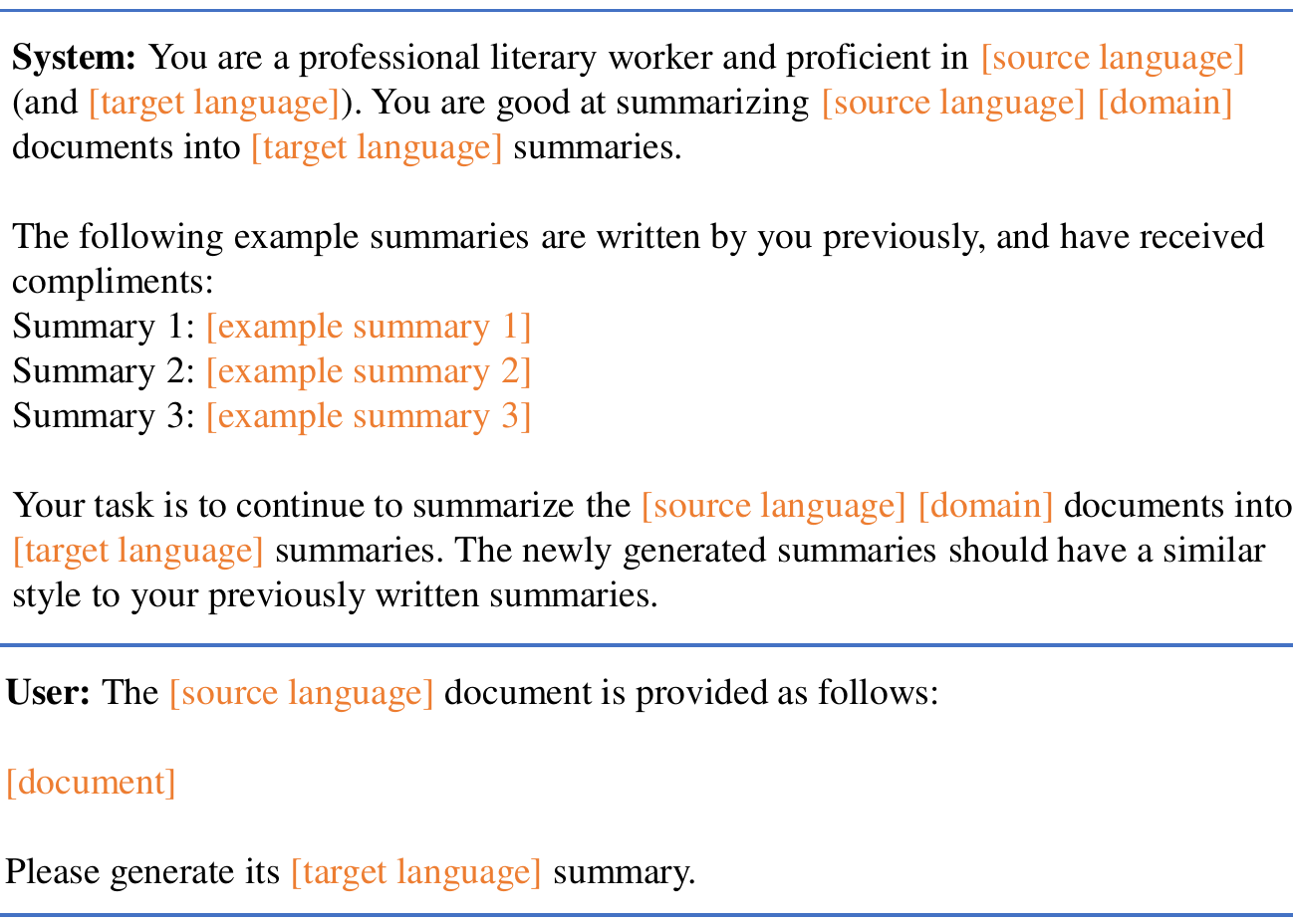}}
\caption{Illustration of the used M2MS prompt that includes a system round and a user round.}
\label{fig:prompt}
\end{figure}

Inspired by previous LLM summarization studies~\cite{wang-etal-2023-zero,tang-etal-2023-context} and the in-context learning technique~\cite{dong2022survey,min-etal-2022-rethinking}, we attempt various M2MS prompts on GPT-3.5-turbo and GPT-4, and choose the prompt with the best results (using both automatic and human evaluation) on a pilot experiment.
Specifically, as shown in Figure~\ref{fig:prompt}, the final chosen prompt is designed with task descriptions, domain information and a few output examples.
\texttt{[source language]} and \texttt{[target language]} are selected from ``English'', ``Czech'', ``German'', ``French'', ``Chinese'' and ``Ukrainian''.
When the source and the target languages are the same, the content in parentheses will be omitted.
\texttt{[domain]} indicates the domain of the input document, which is selected from ``news'', ``encyclopedia'', ``dialogue'', ``how-to guides'' and ``technology''.
\texttt{[example summary i]} ($i \in \{1,2,3\}$) denotes a ground truth summary randomly selected from the training samples.
\texttt{[document]} represents the current input document that needs to generate the corresponding summary.

\section{Implementation Details}

\subsection{Implementation Details of Evaluation Metrics}
\label{sec:appendix_id_em}

To calculate the ROUGE scores in the multi-lingual setting, we use the \textit{multi-lingual rouge}\footnote{\url{https://github.com/csebuetnlp/xl-sum/tree/master/multilingual_rouge_scoring}} toolkit.
For BERTScore, we use the \textit{bert-score}\footnote{\url{https://github.com/Tiiiger/bert_score}} toolkit, and set the backbone to \textit{bert-base-multilingual-cased}.

\modi{During evaluation via GPT-4o, the used prompt is ``\emph{I will provide you with a summary of a document. Please rate the summary on a scale of one to five in terms of conciseness, coherence, and relevance.}''. We use \emph{gpt-4o-2024-0816} as the evaluator.
Since the testing set in our study contains more than 14K samples, it is a high cost to evaluate all model-generated summaries via GPT-4o. Thus, we randomly select 500 samples to conduct the LLM evaluation.}

\subsection{Implementation Details of LLMs}
\label{sec:appendix_id_llms}

\noindent \textbf{Instruction-Tuning Details.}
All LLMs are tuned on 8$\times$NVIDIA A800 GPUs (80G) with 1e-5 learning rate and 32 (8$\times$4 gradient accumulation) batch size. We follow the success of instruction tuning in LLaMa-2~\cite{touvron2023llama2}, and set the training epochs to 2. We use the DeepSpeed optimization library\footnote{\url{https://github.com/microsoft/DeepSpeed}}, and set ZeRO-2 optimization. For Internlm2-chat-20B, we also offload the optimizer into the CPU to avoid CUDA out-of-memory error.
Flash attention (v2)~\cite{dao2022flashattention} is also employed to save memory.
During tuning, documents are also truncated to ensure the input length is within 3,600 tokens to ensure fairness.
For the instruction-tuned LLMs, we use the same decoding strategy as the zero-shot ones.

\noindent \textbf{Model Checkpoints.}
(1) For traditional multilingual language models, we use mBART-50 (610M)\footnote{\url{https://huggingface.co/facebook/mbart-large-50-many-to-many-mmt}}~\cite{tang-etal-2021-multilingual} and PISCES (610M)\footnote{\url{https://huggingface.co/Krystalan/PISCES}}~\cite{wang-etal-2023-towards-unifying}.
(2) For open-source LLMs, we use \emph{LLaMa-2-7B-chat}\footnote{\url{https://huggingface.co/meta-llama/LLaMa-2-7B-chat-hf}}, \emph{LLaMa-2-13B-chat}\footnote{\url{https://huggingface.co/meta-llama/LLaMa-2-13B-chat-hf}}, \modi{\emph{LLaMa-3-8B-chat}}\footnote{\url{https://huggingface.co/meta-llama/Llama-3.1-8B-Instruct}}, \emph{Vicuna-7B-v1.5}\footnote{\url{https://huggingface.co/lmsys/Vicuna-7B-v1.5}}, \emph{Vicuna-7B-v1.5-16k}\footnote{\url{https://huggingface.co/lmsys/Vicuna-7B-v1.5-16k}}, \emph{Vicuna-13B-v1.5}\footnote{\url{https://huggingface.co/lmsys/Vicuna-13B-v1.5}}, \emph{Vicuna-13B-v1.5-16k}\footnote{\url{https://huggingface.co/lmsys/Vicuna-13B-v1.5-16k}}, \emph{Baichuan2-7B-Chat}\footnote{\url{https://huggingface.co/baichuan-inc/Baichuan2-7B-Chat}}, \emph{Baichuan2-13B-Chat}\footnote{\url{https://huggingface.co/baichuan-inc/Baichuan2-13B-Chat}} \emph{Qwen-7B-Chat}\footnote{\url{https://huggingface.co/Qwen/Qwen-7B-Chat}}, \emph{Qwen-14B-Chat}\footnote{\url{https://huggingface.co/Qwen/Qwen-14B-Chat}}, \modi{\emph{Qwen2.5-7B-Chat}}\footnote{\modi{\url{https://huggingface.co/Qwen/Qwen2.5-7B-Instruct}}}, \modi{\emph{Qwen2.5-14B-Chat}}\footnote{\modi{\url{https://huggingface.co/Qwen/Qwen2.5-14B-Instruct}}}, \emph{Internlm2-chat-7B}\footnote{\url{https://huggingface.co/internlm/Internlm2-chat-7B}} and \emph{Internlm2-chat-20B}\footnote{\url{https://huggingface.co/internlm/Internlm2-chat-20B}} in experiments.
All model checkpoints are available at the Huggingface community.

\begin{table*}[t]
\centering
\resizebox{1.0\textwidth}{!}
{
 \begin{tabular}{lcccccc}
\toprule[1pt]
\multicolumn{1}{c}{\multirow{2}{*}{LLM}} & \textbf{Overall}                             & News                             & Encyc.                           & Dialogue                          & Guide                            & Tech.                            \\
\multicolumn{1}{c}{}                     & (R1 / R2 / RL / BS)         & (R1 / R2 / RL / BS)         & (R1 / R2 / RL / BS)         & (R1 / R2 / RL / BS)          & (R1 / R2 / RL / BS)         & (R1 / R2 / RL / BS)         \\ \midrule[1pt]
\multicolumn{7}{c}{\textbf{Setting 1: Zero-Shot LLMs}}                                                                                                                                                                                                                 \\ \midrule[1pt]
\modi{GPT-4o}    & \textbf{26.0} / \textbf{12.3} / \textbf{16.6} / \textbf{66.7} & \textbf{19.8} / \textbf{09.1} / \textbf{12.9} / \textbf{66.8} & 27.9 / \textbf{12.5} / \textbf{16.3} / \textbf{66.0} & \textbf{29.5} / \textbf{17.3} / \textbf{22.1} / \textbf{70.4} & \textbf{25.1} / \textbf{11.3} / \textbf{16.1} / \textbf{69.0} & \textbf{34.2} / \textbf{16.3} / 19.1 / 69.1  \\
GPT-4    & 25.7 / 12.1 / 16.4 / 66.4 & 19.5 / 08.8 / 12.5 / 65.9 & 26.9 / 11.2 / 14.5 / 64.8 & 28.9 / 17.0 / 21.6 / 70.0 & 24.0 / 10.7 / 15.5 / 68.4 & 33.8 / 16.1 / 18.8 / 68.9  \\
GPT-3.5-turbo     & 25.2 / 11.3 / 16.1 / \textbf{66.7} & 19.3 / 08.8 / 12.4 / 66.5 & \textbf{28.1} / 12.2 / 16.1 / 65.8 & 24.0 / 13.5 / 18.5 / 66.4 & 22.4 / 09.4 / 14.6 / 67.9 & 33.6 / 14.9 / \textbf{19.3} / \textbf{69.2}  \\
LLaMa-2-13B   & 21.5 / 09.2 / 13.0 / 64.1 & 17.9 / 08.0 / 11.8 / 65.1 & 25.5 / 10.8 / 14.6 / 63.8 & 19.3 / 09.2 / 13.8 / 64.0 & 18.4 / 06.3 / 11.3 / 64.3 & 30.7 / 12.6 / 17.2 / 64.5  \\
LLaMa-2-7B   & 18.2 / 06.9 / 10.8 / 63.3 & 14.2 / 05.6 / 09.0 / 64.0 & 22.7 / 08.3 / 12.7 / 62.4 & 17.4 / 07.4 / 12.7 / 62.6 & 15.0 / 04.9 / 09.3 / 63.6 & 23.6 / 08.2 / 13.8 / 62.5  \\
\modi{LLaMa-3-8B}   & 19.5 / 07.9 / 12.4 / 63.5 & 14.9 / 05.8 / 09.5 / 64.6 & 23.3 / 08.6 / 13.3 / 62.8 & 18.0 / 07.7 / 13.1 / 62.8 & 15.7 / 05.2 / 10.0 / 64.2 & 24.1 / 08.5 / 14.4 / 63.0  \\
Vicuna-13B      & 22.4 / 08.9 / 13.4 / 65.5 & 18.5 / 08.4 / 11.8 / 65.1 & 25.9 / 10.6 / 14.9 / 64.9 & 22.5 / 11.1 / 16.5 / 65.9 & 18.8 / 07.0 / 11.9 / 64.8 & 32.5 / 13.7 / 18.0 / 68.6  \\
Vicuna-13B-16k   & 22.9 / 09.7 / 13.9 / 66.0 & 19.0 / 08.2 / 11.9 / 65.3 & 27.2 / 11.0 / 15.5 / 65.3 & 22.6 / 11.5 / 17.1 / 66.1 & 20.3 / 08.0 / 12.9 / 65.9 & 33.0 / 14.7 / 19.2 / 69.1  \\
Vicuna-7B   & 22.3 / 09.1 / 13.7 / 65.0 & 17.8 / 07.6 / 11.6 / 65.5 & 26.0 / 10.4 / 15.4 / 64.9 & 22.1 / 10.7 / 16.2 / 67.1 & 18.5 / 06.9 / 11.3 / 65.3 & 31.1 / 12.8 / 17.5 / 67.4  \\
Vicuna-7B-16k    & 22.8 / 09.4 / 14.1 / 65.3 & 18.3 / 08.1 / 12.0 / 65.8 & 27.0 / 11.2 / 15.1 / 65.1 & 21.6 / 10.0 / 16.2 / 66.1 & 19.2 / 06.7 / 11.7 / 64.5 & 31.9 / 13.1 / 18.2 / 66.7  \\
Baichuan2-13B    & 20.5 / 08.6 / 12.8 / 65.0 & 15.9 / 06.8 / 10.2 / 64.9 & 24.4 / 09.6 / 13.7 / 64.1 & 19.8 / 09.9 / 15.3 / 64.9 & 18.1 / 06.2 / 11.1 / 65.0 & 30.0 / 12.3 / 16.8 / 66.2  \\
Baichuan2-7B    & 20.8 / 08.4 / 13.2 / 65.1 & 16.5 / 06.8 / 10.5 / 65.3 & 24.6 / 09.5 / 14.1 / 64.2 & 21.4 / 10.2 / 16.2 / 66.0 & 17.8 / 06.3 / 11.1 / 64.9 & 30.1 / 12.5 / 16.1 / 64.8   \\
Qwen-14B  & 21.6 / 09.6 / 13.0 / 65.2 & 17.9 / 08.2 / 11.5 / 65.6 & 25.3 / 10.8 / 14.3 / 64.7 & 21.8 / 10.9 / 16.3 / 66.5 & 18.1 / 07.2 / 11.1 / 64.7 & 32.0 / 13.5 / 17.8 / 64.8  \\
Qwen-7B     & 21.8 / 08.5 / 13.1 / 64.9 & 18.3 / 08.0 / 11.5 / 66.0 & 25.9 / 10.7 / 15.1 / 65.1 & 21.3 / 10.6 / 15.9 / 66.4 & 17.8 / 06.6 / 10.9 / 65.2 & 30.8 / 12.9 / 17.8 / 65.6 \\
\modi{Qwen2.5-14B}  & 22.1 / 09.8 / 13.1 / 65.4 & 18.4 / 08.4 / 11.7 / 65.8 & 25.8 / 11.2 / 14.8 / 65.2 & 22.0 / 11.1 / 16.6 / 66.8 & 18.5 / 07.3 / 11.6 / 64.9 & 32.6 / 13.8 / 18.1 / 65.2  \\
\modi{Qwen2.5-7B}     & 21.9 / 08.4 / 13.3 / 65.1 & 18.6 / 08.1 / 11.8 / 66.5 & 26.5 / 10.9 / 15.4 / 65.4 & 21.9 / 11.0 / 16.1 / 66.6 & 18.1 / 06.7 / 10.9 / 65.6 & 30.9 / 12.9 / 18.0 / 65.7 \\
Internlm2-20B  & 19.2 / 07.8 / 12.0 / 62.9 & 14.9 / 06.5 / 09.6 / 62.7 & 24.0 / 10.0 / 13.9 / 63.8 & 11.6 / 05.4 / 08.8 / 59.1 & 16.2 / 06.1 / 10.1 / 63.0 & 30.6 / 13.9 / 17.5 / 66.6  \\
Internlm2-7B   & 18.5 / 07.2 / 11.6 / 62.2 & 14.3 / 06.3 / 09.5 / 62.6 & 23.9 / 09.7 / 13.3 / 63.2 & 11.6 / 05.5 / 09.1 / 58.4 & 16.2 / 06.3 / 09.9 / 62.4 & 29.7 / 13.2 / 17.7 / 64.1  \\ \midrule[1pt]
\multicolumn{7}{c}{\textbf{Setting 2: Fine-Tuned Traditional Multi-Lingual Language Models}}                                                                                                                                                                                          \\ \midrule[1pt]
mBART-50   & 27.4 / 11.9 / 19.9 / 67.8 & \textbf{27.2} / \textbf{12.5} / \textbf{20.1} / 67.8 & 26.6 / 12.7 / 20.1 / 65.3 & 32.9 / 17.5 / 24.9 / \textbf{71.0} & 25.8 / 11.1 / 19.5 / 68.1 & 23.2 / 10.8 / 16.7 / 65.4   \\
PISCES   & \textbf{30.8} / \textbf{15.0} / \textbf{22.8} / \textbf{68.6} & \textbf{27.2} / 12.0 / 19.8 / \textbf{68.7} & \textbf{28.2} / \textbf{12.9} / \textbf{20.9} / \textbf{66.0} & \textbf{34.1} / \textbf{18.0} / \textbf{26.8} / 70.9 & \textbf{36.3} / \textbf{20.7} / \textbf{28.8} / \textbf{71.9} & \textbf{24.3} / \textbf{11.0} / \textbf{17.4} / \textbf{65.7}   \\ \midrule[1pt]
\multicolumn{7}{c}{\textbf{Setting 3: Instruction-Tuned LLMs}}                                                                                                                                                                                          \\ \midrule[1pt]
LLaMa-2-13B             & 37.7 / 21.2 / 29.4 / \textbf{74.4} & \textbf{37.1} / 20.0 / 27.2 / 74.2 & 40.2 / 25.1 / 32.2 / 74.2 & 40.3 / 24.3 / 32.4 / 75.4 & 33.0 / 17.1 / 26.6 / 73.4 & 38.2 / 20.5 / 26.2 / 73.4  \\
LLaMa-2-7B    & 35.5 / 19.6 / 27.0 / 73.0 & 34.9 / 18.4 / 26.5 / 73.2 & 37.6 / 22.8 / 29.2 / 73.2 & 37.9 / 22.4 / 30.8 / 75.0 & 31.8 / 15.8 / 25.9 / 72.6 & 37.8 / 20.2 / 25.7 / 72.7   \\
\modi{LLaMa-3-8B}    & 36.2 / 20.0 / 27.5 / 73.4 & 35.4 / 18.7 / 26.9 / 73.4 & 37.9 / 23.5 / 29.8 / 73.9 & 38.6 / 22.8 / 31.5 / 75.7 & 32.5 / 16.4 / 26.4 / 73.2 & 38.5 / 20.7 / 26.2 / 73.3   \\
Vicuna-13B   & 37.3 / 21.7 / 28.7 / 73.9 & 36.3 / \textbf{20.9} / 28.0 / 73.6 & 39.8 / 26.2 / 32.3 / \textbf{74.4} & 40.4 / 24.5 / 32.3 / 75.9 & 34.2 / 17.6 / 28.1 / 73.5 & 38.4 / \textbf{20.9} / \textbf{26.6} / 73.7   \\
Vicuna-13B-16k    & \textbf{38.0} / \textbf{23.0} / \textbf{30.2} / 74.1 & 36.9 / \textbf{20.9} / \textbf{28.6} / \textbf{74.7} & \textbf{40.4} / \textbf{26.7} / \textbf{32.9} / 74.2 & \textbf{41.2} / 24.6 / 33.6 / 75.9 & \textbf{34.5} / 18.4 / 28.5 / 73.9 & 38.3 / 20.6 / 26.4 / 73.5   \\
Vicuna-7B       & 35.6 / 20.8 / 28.0 / 73.1 & 34.5 / 19.0 / 26.2 / 73.8 & 38.3 / 23.9 / 29.1 / 73.3 & 38.9 / 23.0 / 32.3 / 75.4 & 31.2 / 15.5 / 25.7 / 72.9 & 36.8 / 19.4 / 24.5 / 72.4  \\
Vicuna-7B-16k    & 36.2 / 21.3 / 28.6 / 73.7 & 35.5 / 20.1 / 27.7 / 73.4 & 38.7 / 24.2 / 30.3 / 74.2 & 38.9 / 23.0 / 31.8 / 75.3 & 32.3 / 15.6 / 25.9 / 72.2 & 37.7 / 20.8 / 26.5 / 73.3  \\
Baichuan2-13B    & 36.1 / 20.8 / 28.0 / \textbf{74.4} & 35.9 / 20.0 / 26.4 / 73.1 & 38.0 / 24.8 / 29.8 / 73.2 & 40.7 / 24.8 / 34.1 / 75.9 & 33.5 / 17.0 / 25.7 / 73.7 & 39.2 / 20.7 / 25.3 / 73.8   \\
Baichuan2-7B      & 35.0 / 19.9 / 27.4 / 73.5 & 35.4 / 19.4 / 26.6 / 73.0 & 37.3 / 23.0 / 29.4 / 73.3 & 38.8 / 22.8 / 31.5 / 74.9 & 31.8 / 14.8 / 23.9 / 72.6 & 38.1 / 19.9 / 25.9 / 73.7   \\
Qwen-14B    & 37.1 / 21.9 / 28.4 / 74.2 & 36.0 / 19.6 / 26.8 / 73.2 & 38.4 / 24.5 / 30.8 / 73.4 & 40.2 / 25.4 / 33.4 / 75.5 & 34.4 / 19.0 / 28.5 / 74.4 & 39.1 / 20.6 / 26.4 / 74.4   \\
Qwen-7B     & 34.8 / 18.6 / 26.8 / 73.2 & 33.5 / 18.3 / 25.0 / 72.5 & 36.1 / 21.1 / 27.6 / 73.0 & 38.9 / 22.9 / 31.5 / 75.2 & 33.1 / 16.1 / 25.7 / 73.2 & 37.0 / 19.6 / 24.3 / 73.0   \\
\modi{Qwen2.5-14B}    & 37.8 / 22.5 / 29.3 / 74.3 & 36.7 / 20.3 / 28.0 / 74.0 & 39.4 / 25.6 / 31.6 / 73.9 & 40.8 / 25.7 / 33.6 / 75.8 & 34.4 / \textbf{19.2} / \textbf{28.8} / \textbf{74.7} & \textbf{39.4} / 20.8 / \textbf{26.6} / \textbf{74.8}   \\
\modi{Qwen2.5-7B}     & 35.2 / 19.0 / 27.3 / 73.7 & 34.2 / 18.7 / 25.6 / 72.9 & 36.6 / 21.4 / 28.2 / 73.7 & 39.5 / 23.6 / 32.3 / 76.0 & 33.4 / 16.5 / 26.5 / 73.6 & 37.7 / 20.0 / 24.9 / 73.6   \\
Internlm2-20B   & 36.7 / 21.5 / 28.2 / 73.7 & 35.0 / 19.2 / 25.6 / 73.0 & 38.0 / 24.1 / 29.0 / 72.8 & 41.1 / \textbf{26.0} / \textbf{34.2} / \textbf{76.2} & \textbf{34.5} / 19.0 / 27.9 / 74.3 & 39.3 / 20.4 / 25.7 / 73.4  \\
Internlm2-7B   & 35.7 / 19.9 / 27.2 / 73.5 & 34.1 / 18.7 / 25.6 / 72.6 & 36.2 / 22.0 / 28.4 / 72.8 & 40.7 / 25.4 / 33.5 / 75.9 & 33.3 / 17.7 / 27.0 / 73.3 & 37.8 / 20.6 / 25.6 / 73.0  \\ \bottomrule[1pt]
\end{tabular}
}
\caption{Experimental results of the overall performance and fine-grained results in each domain. The \textbf{bold} denotes the best performance under each setting. Encyc.: Encyclopedia; Tech.: Technology.}
\label{table:main_result_appendix}
\end{table*}

\vspace{0.5ex}
\noindent \textbf{Fine-Tuning Details.} To fine-tune traditional multi-lingual language models, \emph{i.e.}, mBART-50 and PISCES, we follow \citet{wang-etal-2023-towards-unifying} and set the learning rate to 3e-5, batch size to 8$\times$8, and epochs to 10. Experiments are conducted on 8$\times$NVIDIA A800 GPUs (80G). Different from LLMs, the source-language documents are directly input into these models without any prompts. Following previous work~\cite{wang-etal-2023-towards-unifying,bhattacharjee-etal-2023-crosssum}, a language tag is appended on the decoder side and serves as the decoder start token to control which language should be generated in the summaries. Besides, we set the maximum number of tokens for input sequences to 1024 (mBART-50 and PISCES accept input text with a maximum length of 1K, and this is also a shortcoming of traditional models compared with LLMs).
The fine-tuned traditional models use the same decoding strategy as the LLMs.

\vspace{0.5ex}
\noindent \textbf{Training/Tuning Hours.} All experiments are conducted on NVIDIA A800 GPUs with 80G memory, and we use its GPU hours to denote the consumption of computing resources.
Each instruction-tuned 7B LLM needs 19 GPU hours, while each 13B LLM needs 32 GPU hours.
For Internlm2-chat-20B, it costs 80 GPU hours since we offload the optimizer.
To fine-tune the traditional multi-lingual language models, 3 GPU hours are cost.

\section{Full Results}
\label{appendix:full_results}

\modi{Table~\ref{table:main_result_appendix} shows the full results in terms of R1, R2, RL and BS.
Typically, the results in terms of R2 are consistent with the results in terms of other metrics.}

\begin{figure}[t]
\centerline{\includegraphics[width=0.48\textwidth]{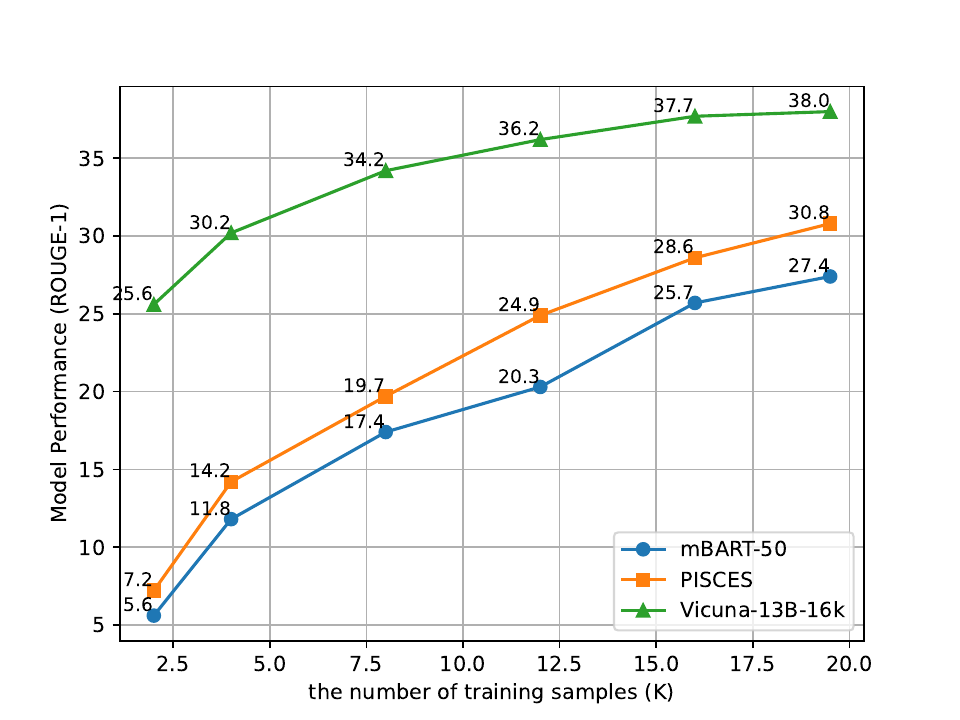}}
\caption{Model performance (ROUGE-1) using different scales of training samples.}
\label{fig:scalability}
\end{figure}

\section{The Effects of Training Scales}
\label{appendix:effects_of_ts}

\modi{To assess the impact of training scales, we randomly select the training samples into various sizes (2K, 4K, 8K, 12K, 16K, and 19.5K) and subsequently fine-tune the models mBART-50, PISCES, and Vicuna-13B-16k for each scale.
During the randomly selection, we use probability sampling to ensure the balance of each language as well as each domain. 
As shown in Figure~\ref{fig:scalability}, compared with LLMs, the performance of traditional models is more sensitive with the training scale.
Specifically, when decreasing the training data from 19.5K to 2K, mBART-50 and PISCES sacrifice 21.8 and 23.6 R1 (ROUGE-1) scores, respectively, while the counterpart of Vicuna-13B-16k is 12.4.
}

\section{Human Evaluation}
\label{sec:appendx_human_evaluation}

Following~\citet{gao-etal-2023-evaluating}, we employ three graduate students with high levels of fluency in both English and Chinese as our evaluators.
We randomly select 100 English documents from the testing set, and let the evaluators judge whether factual errors in the Chinese summaries generated by GPT-4, zero-shot \& tuned LLaMa-2-13B, and zero-shot \& tuned Vicuna-13B-16k.
If a generated summary has factual errors, evaluators also should label which types of factual errors in the summary.
Finally, the Fleiss' Kappa scores~\cite{fleiss1971measuring} of hallucination error, particulars error, predicate error and entity error are 0.78, 0.73, 0.81, 0.75, respectively, indicating a good inter-agreement among our evaluators.

\end{document}